%% file: colm2026_conference.tex
\pdfoutput=1

\documentclass{article}
\usepackage[final]{colm2026_conference}

\usepackage{microtype}
\usepackage[labelfont=bf,font=small]{caption}
\usepackage{subcaption}
\usepackage{float}
\usepackage[section]{placeins}

\usepackage{hyperref}
\usepackage{url}
\usepackage{booktabs}
\usepackage{multirow}
\usepackage{graphicx}
\usepackage{adjustbox}
\usepackage[most]{tcolorbox}
\usepackage{colortbl}
\usepackage[T1]{fontenc}
\usepackage{xspace}
\usepackage{fontawesome5}

\usepackage{lineno}
\usepackage{wrapfig}
\usepackage[normalem]{ulem}
\usepackage[nameinlink]{cleveref}

\definecolor{darkblue}{rgb}{0, 0, 0.5}
\hypersetup{colorlinks=true, citecolor=darkblue, linkcolor=darkblue, urlcolor=darkblue}

\title{Reasoning with Image Generation}

\author{Nishad Singhi$^{*\dagger}$, Hector Garcia Rodriguez$^{*\dagger}$, Aditya Arora$^\dagger$, \\
\textbf{Marcus Rohrbach, Anna Rohrbach} \\
Technical University of Darmstadt \& hessian.AI
}

\newcommand{\ours}{\textsc{ReImaGin\xspace}}
\newcommand{\oursLong}{\underline{Re}asoning with \underline{Ima}ge \underline{G}enerat\underline{i}o\underline{n}}
\newcommand{\gr}[1]{\textcolor{gray!60}{#1}}
\newtcblisting{promptfloatlisting}[1][]{%
  enhanced,
  colback=white,
  colframe=black!80,
  boxrule=1pt,
  arc=2.5mm,
  left=2mm,
  right=2mm,
  top=0.5mm,
  bottom=0.5mm,
  title filled=false,
  colbacktitle=white,
  coltitle=black,
  fonttitle=\bfseries,
  listing only,
  listing options={
    basicstyle=\ttfamily\footnotesize,
    breaklines=true,
    breakatwhitespace=true,
    columns=fullflexible,
    keepspaces=true,
    showstringspaces=false
  },
  #1
}
\newcommand{\promptfloatmaxheight}{0.74\textheight}
\newenvironment{fitpromptbox}{%
  \begin{adjustbox}{max totalsize={\linewidth}{\promptfloatmaxheight},center}
  \begin{minipage}{\linewidth}
}{%
  \end{minipage}
  \end{adjustbox}
}

\begin{document}

\ifcolmsubmission
\linenumbers
\fi

\newcommand\blfootnote[1]{%
  \begingroup
  \renewcommand\thefootnote{}\footnote{#1}%
  \addtocounter{footnote}{-1}%
  \endgroup
}

\maketitle
\blfootnote{$^*$ Equal contribution \quad $^\dagger$ Core contributor \quad \href{https://github.com/multimodal-ai-lab/reimagin}{\faGithub\, Code}}

\input{files/abstract}

\vspace{-0.8em}
\begin{figure}[H]
    \centering
    \includegraphics[width=0.95\textwidth]{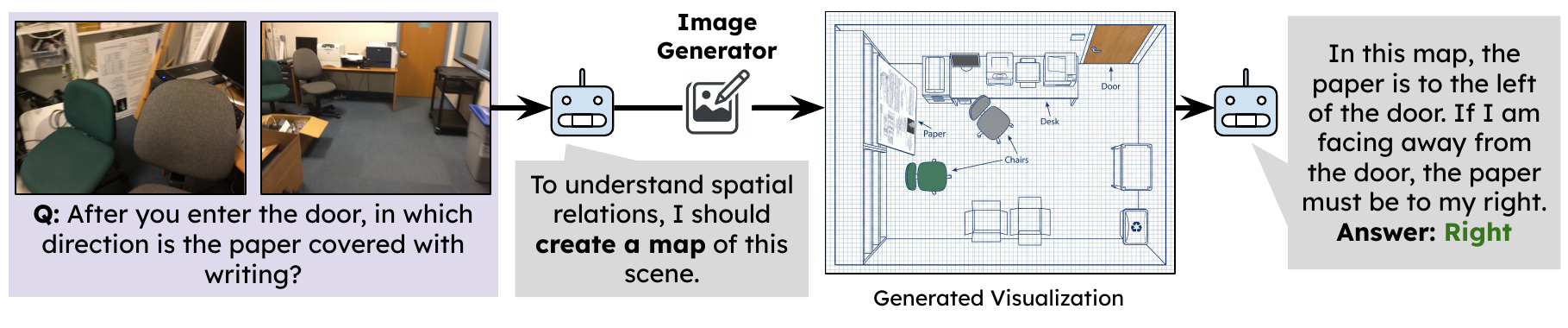}\\[0.3em]
    \vspace{-0.05cm}
    \includegraphics[width=0.67\textwidth]{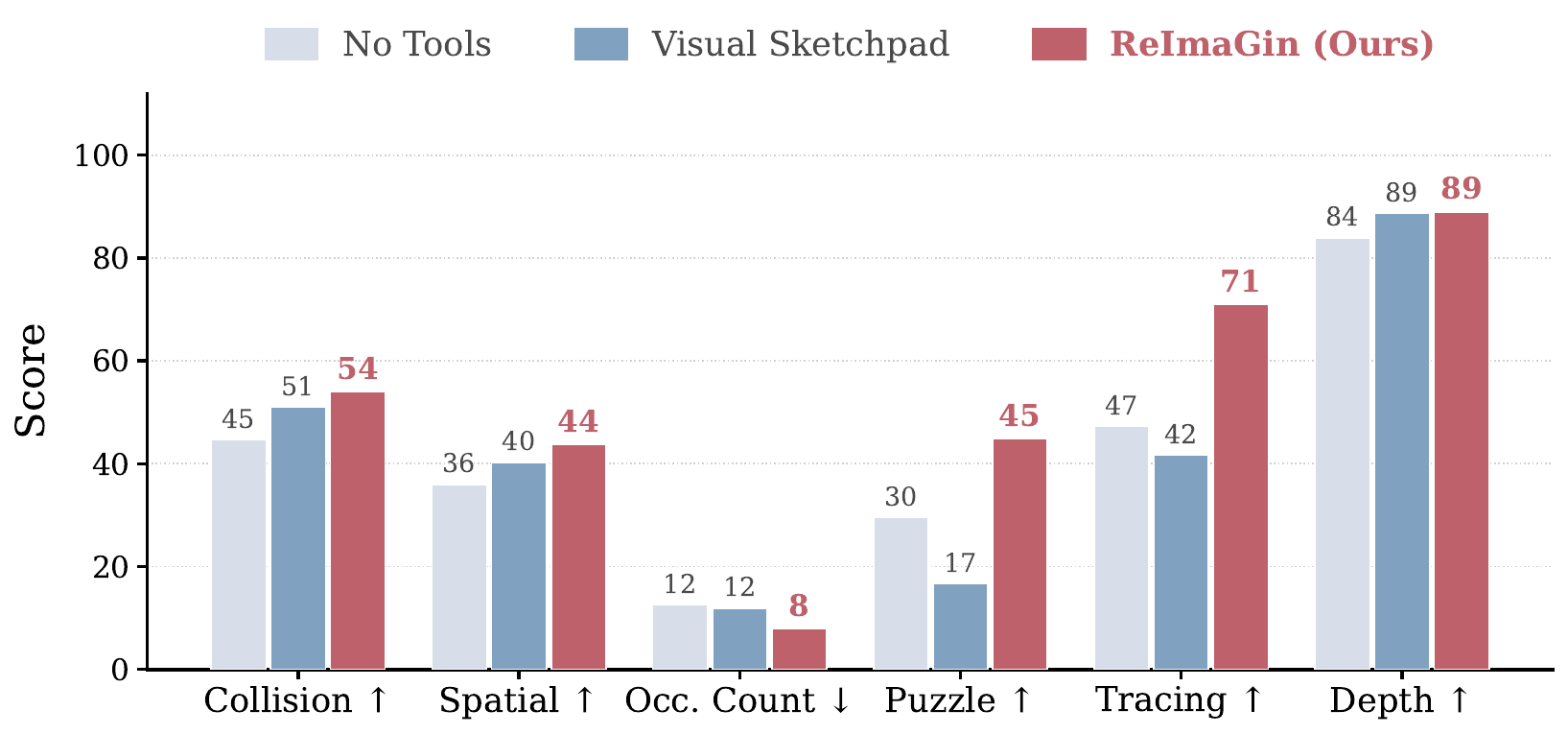}
    \caption{\textbf{Top.} We propose a multimodal reasoning framework \ours{} that solves complex visual tasks by exploiting image generation models as an intermediate state of reasoning. This showcases the spatial reasoning task from the MMSI~\citep{yang2025mmsi} benchmark. Answering this spatial question requires fusing information from two views of the same room, since neither view contains both the door and the paper. \ours\ addresses this by invoking a generative model to produce a top-down map that consolidates objects from both views into a single frame, and then reasoning over this unified visualization. \textbf{Bottom.} Performance with GPT-5: \ours{} consistently outperforms the no tools and the specialist-tool baseline Visual Sketchpad~\citep{hu2024visual}.}
    \label{fig:teaser}
\end{figure}

\input{files/intro}

\input{files/related_work}
\input{files/method}

\input{files/experiments}

\input{files/results}
\input{files/conclusion}

\clearpage
\appendix
\begin{center}
  \textbf{\Large Appendix}
\end{center}
\input{files/appendix}

\end{document}

%% file: files/abstract.tex
\begin{abstract}
Chain-of-thought reasoning has revolutionized natural language processing by enabling large language models (LLMs) to decompose problems into intermediate steps before answering. 
Yet confining reasoning to the textual domain presents limitations for tasks requiring direct manipulation of visual representations. Recent efforts augment multimodal LLMs with external visual expert tools such as depth estimation or object detection modules, but these remain fundamentally limited by their reliance on narrow, rigid operations that cannot flexibly generate or transform visual content. 
We propose \ours, which leverages image generation models as a flexible visual reasoning mechanism for multimodal LLMs: 
unlike fixed-function tools, they accept natural language commands and can perform open-ended visual operations, like removing an occlusion or generating a floorplan from multiple disjoint views of a room.
Across six diverse visual reasoning tasks including multi-view spatial reasoning and collision prediction, \ours{} consistently outperforms both text-only reasoning and specialist vision-tool baselines, with gains of up to 24\%, demonstrating the advantage of flexible, generative visual reasoning.
\end{abstract}

%% file: files/intro.tex
\section{Introduction}

Chain-of-Thought (CoT;~\citet{wei2022chain}) reasoning has emerged as a paradigm shift in natural language processing, significantly enhancing the capabilities of Large Language Models (LLMs) by enabling them to generate intermediate reasoning rationales. This success has naturally extended to the multimodal domain, driving impressive performance gains in tasks such as Visual Question Answering (VQA)~\citep{zhang2023multimodal, lu2022learn, alayrac2022flamingo, liu2023visual, dai2023instructblip}. However, the vast majority of existing multimodal frameworks restrict the ``reasoning'' process to the textual domain, essentially describing visual inputs in words and processing them logically~\citep{alayrac2022flamingo, li2023blip2, liu2023visual, dai2023instructblip}. While effective for some tasks, this text-centric approach is ill-suited for tasks that demand spatial or physical intuition, such as visualizing the removal of occlusions or novel view synthesis.

To overcome the limitations of purely textual reasoning, recent works have begun augmenting Multimodal LLMs (MLLMs) with external visual tools, such as modules for cropping, depth estimation, and object detection~\citep{hu2024visual, fu2025refocus}. These frameworks allow a model to execute basic visual operations to support its reasoning process. However, this approach suffers from two limitations. The primary limitation is the rigidity of the tools provided: these frameworks rely on pre-defined modules that can identify a bounding box or segment an object, but are unable to perform flexible, generative, or complex transformations (e.g., imagining an alternative view of a scene). Consequently, the reasoning process remains limited by the static and narrow nature of the underlying toolset. A second limitation is that models are typically taught to use these tools via handcrafted in-context examples, adding manual effort and limiting generalization to new tasks.

To address the narrow capabilities of traditional visual tools, we explore a new framework: \emph{reasoning with image generation}. Unlike fixed-function tools that are limited to a predetermined set of operations (e.g., segmentation, depth estimation), recent generative models are trained to \emph{follow instructions in natural language}, enabling them to perform a vast and open-ended range of \emph{visual operations that can be expressed linguistically}~\citep{2025nanobananapro, flux-2-2025, wu2025qwenimagetechnicalreport}.
This flexibility is transformative: a single generative model integrates the capabilities of many specialist vision tools (e.g., segmentation and depth estimation) while also producing arbitrary visualizations and image transformations that were previously out of reach, such as alternative views or a blueprint of a room, without any task-specific specialization. 
Crucially, the capabilities of this approach scale directly with advances in visual generation: as generative models become more capable, the range of ways the agent can reason visually expands accordingly.

We instantiate this idea as \ours, a multimodal agent that interleaves textual chain-of-thought with calls to an instruction-tuned image generation model. The agent is equipped with a free-form \texttt{generate\_image} tool that can be invoked at any reasoning step with a natural language prompt, returning a visual intermediate directly into the agent's context (see Figure~\ref{fig:teaser}). For instance, on a collision prediction task, the agent might call \texttt{generate\_image} to draw a trajectory line from the moving object, then inspect the resulting image to identify which object the line first intersects (Figure~\ref{fig:qualitative}). Unlike specialist vision tools, which expose a fixed set of operations, the same \texttt{generate\_image} tool can perform a broad range of visual transformations, based on the natural-language prompt it receives. %

Beyond the core framework, we also investigate a question that the flexibility of generative tools raises. Because \texttt{generate\_image} accepts arbitrary natural-language instructions, the agent can perform a vast range of transformations; the question is which transformation actually helps for a given task. As noted earlier, standard practice is to specify the transformation through handcrafted in-context examples that demonstrate the desired strategy (e.g., generate a depth map for depth reasoning). This requires manual effort per task and limits generalization to new tasks. As a complementary exploration, we ask whether such strategies can be discovered automatically. Since the strategy is conveyed to the agent through its prompt, discovering a strategy reduces to optimizing the prompt: we instantiate an iterative loop in which a proposal model generates candidate prompts, evaluates them on a small development set, and refines based on observed successes and failures. %

We evaluate \ours\ on six diverse visual reasoning tasks spanning depth perception~\citep{fu2024blink}, visual puzzle completion~\citep{zhou2025visualizing}, counting under partial occlusion~\citep{pothiraj2025capture}, collision prediction~\citep{wang2025spatial457}, multi-view spatial reasoning~\citep{yang2025mmsi}, and a path tracing task that we introduce. The same image generation model serves as the tool across all tasks, performing transformations ranging from depth map generation to occlusion removal and floorplan synthesis, without any task-specific specialization. Across tasks and MLLMs, \ours\ with handcrafted strategies consistently improves over both text-only reasoning and reasoning with fixed visual tools, specifically Visual Sketchpad~\citep{hu2024visual}, by up to 24\% on path tracing and 37\% relative on counting under partial occlusion. Further, automatically discovered strategies often resemble those a human would design (e.g., drawing an arrow from the front of a moving object to predict collisions), suggesting that MLLMs have useful priors about which visual transformations aid a given task. These discovered strategies recover most of the gains of handcrafted ones without any human-crafted examples, though a gap remains on tasks that benefit from more elaborate visual reasoning policies.

Our contributions are three-fold:
\begin{enumerate}
  \item We propose \ours, a multimodal agent framework that equips an MLLM with a free-form image generation tool. Unlike prior work that relies on fixed specialist modules, the generative tool supports a far wider and more flexible range of visual transformations.
  \item Using this same generalist tool across six diverse visual reasoning tasks, we show that \ours\ consistently improves over text-only reasoning and reasoning with expert image tools (i.e.\ Visual Sketchpad).
  \item We additionally show that effective visual reasoning strategies can be discovered automatically via prompt optimization, recovering most of the gains of handcrafted strategies and often resembling them, which suggests MLLMs have useful priors about which visual transformations aid a given task.
\end{enumerate}

%% file: files/related_work.tex
\section{Related Work}

Prior work equips multimodal LLMs with \emph{fixed specialist vision tools} such as crop, detection, segmentation, or depth estimation~\citep{hu2024visual,fu2025refocus,wang2025pixelreasoner,zheng2025deepeyes}, but each tool performs a narrow, pre-specified transformation, so the agent can only apply manipulations implemented in advance. A separate line of work \emph{iteratively refines generated images}~\citep{yang2024idea2img,guo2025can,khan2025tir,wan2025maestro}, where the image is the final object being optimized rather than an intermediate for a downstream question. Closest to us, recent methods \emph{reason with generative or latent visual models}~\citep{li2025imaginereasoningspacemultimodal,xu2026visualplanningletsthink,he2025diffthinkergenerativemultimodalreasoning,gu2025thinkmorph,yang2026machine,qin2025chain}, but they typically require task-specific training, target a single narrow domain, or reason in non-interpretable latent space. In contrast, \ours\ is training-free and modular: it invokes a single instruction-following image generator for open-ended transformations, carries out its visual reasoning in interpretable pixel space, and is evaluated across six diverse visual reasoning tasks. We provide an extended discussion in Appendix~\ref{app:related_work}.

%% file: files/method.tex
\section{\ours{} - \oursLong}
\label{sec:method}

\begin{figure*}[t]
    \centering
    \includegraphics[width=\textwidth]{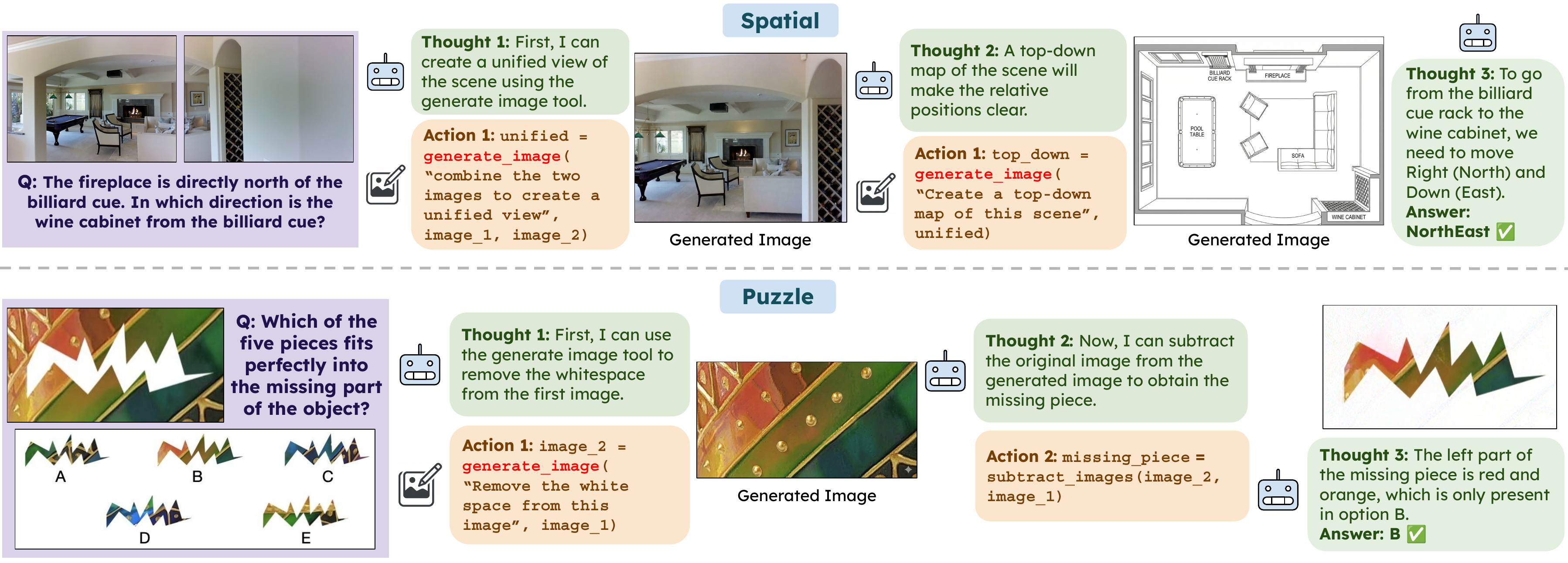}
    \caption{\textbf{Overview of \ours{}}. At each turn, the agent determines the next action, e.g., to generate an image necessary for the reasoning process; that image is then passed on and used in the next turn. Here, we showcase the spatial reasoning task from~\citet{yang2025mmsi} (Top) and the puzzle task from~\citet{zhou2025visualizing} (Bottom).}
    \label{fig:method}
\end{figure*}

We present \ours, a multimodal agent framework that uses instruction-following generative models to produce visualizations within an iterative reasoning loop. The agent can use the generative model by calling the \texttt{generate\_image} function, in addition to standard programmatic image tools (e.g., crop, overlay, subtract, and libraries like \texttt{numpy}), within a python environment. 
Unlike prior tool-augmented approaches, such as Visual Sketchpad \citep{hu2024visual}, which depend on a fixed suite of specialist vision models (e.g., depth estimator, segmenter, object detector) each confined to its own narrow task, \ours\ leverages an instruction-following generative model that can flexibly perform a wide range of operations through natural language. 
Figure~\ref{fig:method} (Appendix~\ref{app:overview}) provides an overview.

Let $\mathcal{Q}$ denote a natural-language query and $\mathcal{V} = \{v_1, \dots, v_n\}$ a set of input images. The goal is to produce an answer $a^*$ by reasoning jointly over $\mathcal{Q}$ and $\mathcal{V}$. At each turn $t$, the agent maintains a context
\[
    C_t = \bigl(\mathcal{Q},\; \mathcal{V},\; h_1, \dots, h_{t-1}\bigr),
\]
where $h_i = (\theta_i, \alpha_i, o_i)$ is the history of the $i$-th turn: a chain-of-thought rationale $\theta_i$, an executable action $\alpha_i$, and the action's output $o_i$.

\subsection{Agent Architecture}\label{sec:agent}
Our agent architecture builds upon the ReAct framework~\citep{yao2022react}, illustrated in Figure~\ref{fig:overview} (Appendix~\ref{app:overview}). We describe each component below, and provide an example in Appendix~\ref{app:worked_ex}.

\textbf{Prompt.} At each turn, an MLLM receives a prompt consisting of three components: tool descriptions that define the available tools, in-context examples that illustrate how to apply visual reasoning to solve the task, and the context $C_t$ comprising the query, input images, and prior turn history.

\textbf{Reasoning step.} At each turn $t$ the MLLM receives $C_t$ and produces a chain-of-thought rationale $\theta_t$ that interprets the previous context and reasons about the next action.

\textbf{Action step.} Based on $\theta_t$, the agent emits a Python program $\alpha_t$ that is executed at runtime. The program may call programmatic image utilities (e.g., \texttt{crop}, \texttt{overlay}), the generative tool \texttt{generate\_image}, and general-purpose code (e.g., \texttt{numpy} for pixel-level analysis). We retain the cheap, deterministic programmatic utilities alongside the generative tool for simple transformations; some tasks compose both, e.g., the puzzle task (Figure~\ref{fig:method}) uses \texttt{generate\_image} to remove whitespace and \texttt{subtract\_images} to isolate the missing piece.

\textbf{Image generation tool.} The \texttt{generate\_image} tool accepts a text prompt and, optionally, one or more reference images, which are passed to an instruction-following image generation model that returns a new synthesized image. %

\textbf{Context update and termination.} The outputs of $\alpha_t$ (including any generated images) are appended to the agent's context as a new user message before the next turn. The loop continues until the agent appends \textsc{Terminate} to its rationale $\theta_t$, at which point the final answer is parsed from the preceding text.

\textbf{Improving Image Generation via Test-Time Scaling.} Image generation is stochastic, and variance grows with task difficulty: simple edits such as occlusion removal are consistent across samples, whereas harder transformations such as floorplan creation can yield very different layouts. Since reasoning accuracy depends on the faithfulness of the generated image, we exploit this stochasticity as test-time scaling: \texttt{generate\_image} samples $N$ candidates independently and uses an MLLM to select the one that most faithfully satisfies the prompt~\citep{karthik2023if}. From the agent's perspective the interface is unchanged---it always receives a single image back. Figure~\ref{fig:test_time_scaling} (Appendix~\ref{app:test_time_scaling}) illustrates this.

Whereas prior test-time scaling draws multiple text chain-of-thought traces and aggregates them via majority voting or best-of-$N$~\citep{brown2024large,singhisolve,snell2024scaling}, we instead sample a single reasoning trace and spend the extra compute on improving the generated images. Multiple CoTs could be sampled for additional test-time compute.

\subsection{Automated Discovery of Visual Reasoning Strategies}\label{sec:discovery}

An important element in our framework is deciding which visual transformations the agent uses to benefit reasoning.
The potential of visual reasoning is maximized when a good solution strategy is used.
For example, in the path tracing task, dashed lines can confuse MLLMs.
Realizing that, e.g., making the lines solid significantly simplifies the reasoning, can be very important.
Prior work \citep{hu2024visual} relies exclusively on hand-crafted strategies for how and when to use the various individual tools in order to help answer the question. 
While effective, hand-crafted strategies require task-specific human effort and reintroduce a human-in-the-loop bottleneck.

To reduce this bottleneck, we ask whether effective visual reasoning strategies can instead be discovered automatically. 
Since the strategy is conveyed to the agent through its prompt (i.e., through tool definitions and in-context examples that demonstrate when and how to invoke each tool), discovering a strategy reduces to finding a prompt that induces effective tool use. 
We therefore formulate automated strategy discovery as an iterative search over agent prompts. 
Given a task $\tau$ with training split $D_{\mathrm{train}}$ and development split $D_{\mathrm{dev}}$, the goal is to obtain a prompt $p^*$ that induces effective visual reasoning for a fixed reasoning agent $A_R$. 
The search is initialized from the baseline \emph{No Strategy} prompt, which includes the standard tool definitions but no task-specific visual reasoning policy. 
At round $r$, the search maintains a set of candidate prompts $\mathcal{P}_r$. 
Each round alternates between two stages: \emph{candidate prompt evaluation} and \emph{candidate prompt proposal}.
In the \emph{candidate prompt evaluation} stage, we run $A_R$ with each prompt $p \in \mathcal{P}_r$ on both $D_{\mathrm{train}}$ and $D_{\mathrm{dev}}$.
Ground-truth answers are used to compute the task metric and identify successful and failed trajectories.
Examples of failure and successful attempts on the training split are seen during the proposal stage to refine the strategies.
The development score $J(p)$ is used for selection of the top strategies. 
In the \emph{candidate prompt proposal} stage, a proposal agent $A_P$ receives the retained prompts together with selected successful and failed trajectories drawn from the training split $D_{\mathrm{train}}$, and proposes new prompts intended to improve the tool use.

The best prompt across rounds is returned as $p^*$.
Throughout the search, $A_R$ and the image generator remain fixed; only the agent prompt is optimized.
We use the same guidance for candidate prompt proposal for all tasks, as shown in Figure~\ref{fig:app_runtime_meta_prompt_logged}.
Notably, the optimization receives no human-authored task-specific strategy or reasoning trace.

%% file: files/experiments.tex
\section{Experimental Setup}
\label{sec:experiments}

\textbf{Tasks.} We evaluate on six tasks spanning visual reasoning problems that require spatial understanding and visual transformations.
\textbf{Depth Reasoning} (BLINK; ~\citet{fu2024blink}) tests relative depth perception: given an image with two marked points, the model must predict which point is closer to the camera.
\textbf{Puzzle Completion} (MIRA; ~\citet{zhou2025visualizing}) presents an image with a missing region alongside five candidate pieces, and the model must identify which piece fits the gap perfectly.
\textbf{Occlusion Counting} (CAPTURe; ~\citet{pothiraj2025capture}) shows multiple instances of an object category, some occluded by a black square, and the model must count them all, including the hidden ones.
\textbf{Collision Prediction} (Spatial457; \citet{wang2025spatial457}) presents overhead-view scenes and asks which object a target would collide with if it moved forward or backward.
\textbf{Spatial Reasoning}  (MMSI; ~\citep{yang2025mmsi}) presents two partially overlapping views of an indoor scene and asks about the relative positions of objects or regions across them; we use the Pos (Obj-Obj) and Pos (Obj-Reg) splits.
\textbf{Path Tracing} is a task we introduce: each image contains four numbers (1--4) and four letters (A--D) at random locations, each number connected to one letter by a dashed line, and the model must identify which letter each number connects to (Figure~\ref{fig:qualitative}).
Additional details about each task are provided in Appendix~\ref{app:benchmarks}.

\textbf{Models.} As the MLLM backbone we use Gemini-3.1-Pro~\citep{gemini2025gemini3} and GPT-5~\citep{openai2025gpt5systemcard} as proprietary models, and Qwen-3.5-27B~\citep{qwen35blog} as an open-weights alternative.
As the visual generative tool, our primary model is Nano-Banana-Pro (Gemini-3-Pro-Image; ~\citet{2025nanobananapro}), a state-of-the-art instruction-tuned image generation model. We additionally experiment with two open-weights generative models: FLUX.2 [dev]~\citep{flux-2-2025} and Qwen-Image-Edit-2511~\citep{wu2025qwenimagetechnicalreport}.
For image-generation test-time scaling, we use Gemini-3.1-Pro as the selector.

\textbf{Baselines.} We compare against two baselines.
\textbf{No Tools} is a vanilla MLLM that answers directly from the input query and images, without access to any tools or code execution.
\textbf{Visual Sketchpad}~\citep{hu2024visual} augments the MLLM with a fixed toolbox of specialist vision modules and programmatic manipulation tools, but does not include any open-ended generative visual capability.
On the spatial task (MMSI), \ours\ samples a single reasoning trace; each image generation call samples $10$ candidate images and a Gemini-3.1-Pro selector keeps one (Section~\ref{sec:agent}).

\textbf{Metrics.} All tasks use multiple-choice accuracy, except Occlusion Counting, for which we report symmetric mean absolute percentage error (sMAPE) ~\citep{pothiraj2025capture} %
, $\mathrm{sMAPE}=100 (\lvert y-\hat y\rvert)/( y + \hat y)$
, where $y$ and $\hat y$ are the ground-truth and predicted counts. We report metrics averaged over 3 seeds, along with the standard error across these runs.

All prompts used in our experiments are provided in Appendix~\ref{app:prompts}.

\textbf{Automated Discovery of Visual Reasoning Strategies.} Here, we compare ``Handcrafted Strategy'', using the default human-designed visual reasoning policy, ``No Strategy'' (employs image generation without task-specific guidance), and our ``Automatic Strategy'', using a policy discovered from development feedback. 
We perform visual reasoning strategy discovery with at most $|D_{\mathrm{train}}| = 50$ and $|D_{\mathrm{dev}}| = 50$ samples, less on MMSI due to dataset size limitations.
We run the search for $R = 4$ rounds and create 5 candidate prompt proposals in each round. 
In practice, both the proposal agent $A_P$ and the reasoning agent $A_R$ are instantiated with the same model. In our experiments, we used Gemini-3.1-Pro or Qwen-3.5-27B.
The proposal context retains the top 2 prompts found so far, and each included prompt context block contributes 2 incorrect and 2 correct examples from $D_{\mathrm{train}}$, as well as other success and failure examples from the last round.
We implement the strategy-discovery loop using Opik \citep{cometml2024opik}.
Full prompts and details on the one-time discovery cost of approximately \$$214$ per task are deferred to Appendix~\ref{app:strategy_prompts}.

%% file: files/results.tex
\section{Results}

\subsection{Main Results}

\begin{table*}[t]
  \centering
  \small
  \resizebox{\linewidth}{!}{
  \begin{tabular}{@{}ll cccccc@{}}
  \toprule
  \textbf{Model} & \textbf{Approach} & \shortstack{\textbf{Puzzle $\uparrow$} \\ {\scriptsize (MIRA)}} & \shortstack{\textbf{Occ.\ Count $\downarrow$} \\ {\scriptsize (CAPTURe)}} & \shortstack{\textbf{Collision $\uparrow$} \\ {\scriptsize (Spatial457)}} & \shortstack{\textbf{Spatial $\uparrow$} \\ {\scriptsize (MMSI)}} & \shortstack{\textbf{Tracing $\uparrow$} \\ {\scriptsize (Ours)}} & \shortstack{\textbf{Depth $\uparrow$} \\ {\scriptsize (BLINK)}} \\
  \midrule
  \multirow{3}{*}{Gemini-3.1-Pro}
    & No Tools          & $37.2_{\pm 2.6}$  & $\underline{8.9}_{\pm 0.4}$  & $60.3_{\pm 0.9}$  &  $49.0_{\pm 0.6}$  &  $70.7_{\pm 1.2}$  & $91.7_{\pm 0.3}$ \\
    & Sketchpad         & $\underline{38.5}_{\pm 8.9}$  &  $9.1_{\pm 0.6}$  &  $\underline{66.7}_{\pm 1.8}$  &  $\underline{55.7}_{\pm 3.5}$  &  $\underline{75.7}_{\pm 2.6}$  &  $\textbf{98.9}_{\pm 0.3}$ \\
    \rowcolor{blue!10} \cellcolor{white} & \ours             & $\textbf{42.3}_{\pm 2.2}$  & $\textbf{7.1}_{\pm 0.6}$  & $\textbf{69.7}_{\pm 1.2}$  & $\textbf{59.0}_{\pm 0.6}$  &  $\textbf{91.0}_{\pm 1.7}$  & $\underline{94.6}_{\pm 1.5}$ \\
  \midrule
  \multirow{3}{*}{GPT-5}
    & No Tools          & $\underline{29.5}_{\pm 3.4}$  & $12.5_{\pm 0.6}$  & $44.7_{\pm 1.9}$  &  $36.0_{\pm 2.0}$  &  $\underline{47.3}_{\pm 1.5}$  & $83.9_{\pm 1.6}$ \\
    & Sketchpad         & $16.7_{\pm 1.3}$  &  $\underline{11.9}_{\pm 0.3}$  &  $\underline{51.0}_{\pm 2.1}$  &  $\underline{40.3}_{\pm 2.4}$  & $41.7_{\pm 1.7}$  &  $\underline{88.7}_{\pm 0.5}$ \\
    \rowcolor{blue!10} \cellcolor{white} & \ours             & $\textbf{44.9}_{\pm 5.6}$  &  $\textbf{7.9}_{\pm 0.3}$  & $\textbf{54.0}_{\pm 1.5}$  &  $\textbf{43.7}_{\pm 3.2}$  &  $\textbf{71.0}_{\pm 2.5}$  &  $\textbf{89.0}_{\pm 1.5}$ \\
  \midrule
  \multirow{3}{*}{Qwen-3.5-27B}
    & No Tools          & $\underline{34.6}_{\pm 2.2}$  &  $12.1_{\pm 0.3}$  &  $\underline{51.7}_{\pm 1.5}$  &  $42.0_{\pm 1.5}$  & $\underline{35.7}_{\pm 1.2}$  &  $85.5_{\pm 2.0}$ \\
    & Sketchpad         & $25.6_{\pm 2.6}$  & $\underline{8.7}_{\pm 0.0}$  &  $48.3_{\pm 1.8}$  &  $\underline{43.0}_{\pm 0.6}$  &  $28.0_{\pm 4.5}$  &  $\underline{89.5}_{\pm 1.6}$ \\
    \rowcolor{blue!10} \cellcolor{white} & \ours             & $\textbf{46.2}_{\pm 2.2}$  &  $\textbf{6.4}_{\pm 0.6}$  & $\textbf{61.3}_{\pm 2.3}$  &  $\textbf{48.0}_{\pm 4.0}$  &  $\textbf{76.0}_{\pm 2.5}$  &  $\textbf{90.1}_{\pm 1.8}$ \\
  \bottomrule
  \end{tabular}
  }
  \caption{\textbf{Results across six visual reasoning tasks.} We compare \ours\ to No Tools and Visual Sketchpad~\citep{hu2024visual} using different MLLMs.  On MMSI, \ours\ keeps the best of $10$ sampled images per image generation call. Occ.\ Count uses sMAPE (lower is better), others accuracy.}
  \label{tab:main}
  \vspace{-15pt}
  \end{table*}
  
In these experiments we rely on the same human-defined in-context strategy examples for all models, for fair comparison.
Across all three models (Table~\ref{tab:main}), \ours\ consistently outperforms both the No Tools and Sketchpad baselines on the majority of tasks. On tasks such as puzzle completion, occlusion counting, and path tracing, Visual Sketchpad often fails to improve over No Tools or even degrades performance (e.g., 16.7\% vs.\ 29.5\% on puzzle completion with GPT-5), as these tasks require generative capability that specialist vision models lack. This underscores the importance of flexible, open-ended visual transformations over fixed specialist tools. The one exception is depth reasoning on Gemini-3.1-Pro, where \ours\ improves over No Tools (94.6\% vs.\ 91.7\%) but trails Visual Sketchpad (98.9\%), which benefits from a dedicated depth estimation specialist.
Since the baselines also differ from \ours\ in their prompts, we provide a further ablation where we ablate only the generative tool while keeping everything else fixed. This degrades performance on all six tasks (Appendix~\ref{app:no_generate}), confirming that the gains come from the generative capability itself.

\begin{wrapfigure}{r}{0.46\textwidth}
  \centering
  \includegraphics[width=0.44\textwidth]{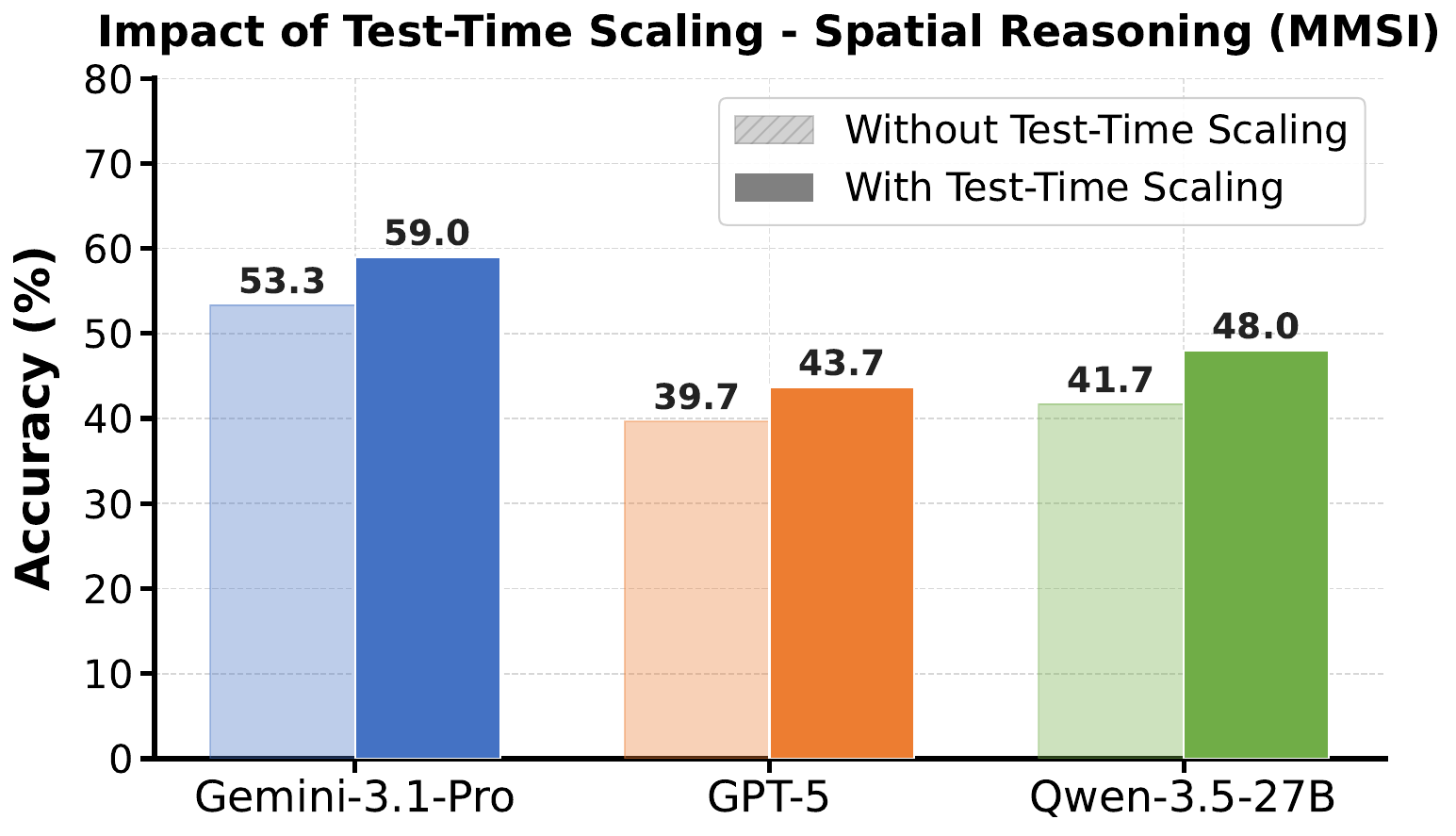}
  \caption{\textbf{Effect of test-time scaling on spatial reasoning across models.} Sampling multiple candidate generations improves accuracy for all three MLLMs.}
  \label{fig:tts_bar_chart}
\end{wrapfigure}
\vspace{-0.05pt}
\textbf{Impact of Test-time Scaling.} For spatial reasoning (MMSI), \ours\ first stitches the two input images into a unified view, then generates a top-down perspective of the scene to make spatial relationships explicit (Figure~\ref{fig:method}, top). This task poses a particular challenge for the generative model: the required geometric operations are difficult, and the generated layouts vary significantly across samples (as shown in Figure~\ref{fig:test_time_scaling_result}, Appendix~\ref{app:test_time_scaling}). To address this, we leverage test-time scaling (as defined in Section~\ref{sec:agent}) with $N=10$ samples, selecting the most faithful generation rather than relying on a single output. Figure~\ref{fig:tts_bar_chart} illustrates the impact across models: without test-time scaling, \ours\ achieves 53.3\% on Gemini-3.1-Pro and 41.7\% on Qwen-3.5-27B. With test-time scaling, performance improves substantially to 59.0\% and 48.0\% respectively. Importantly, test-time scaling only affects the generated images and leaves the rest of the reasoning process unchanged, demonstrating that more faithful visualizations directly translate to better downstream performance.  Giving the baselines extra test-time compute via majority voting over $20$ answers narrows the gap for Gemini-3.1-Pro and closes it for Qwen-3.5-27B, but not for GPT-5 (Appendix~\ref{app:baseline_voting}). The other five tasks show little variance across generated samples (Figures~\ref{fig:test_time_scaling_mira} and ~\ref{fig:test_time_scaling_capture_occlusion}, Appendix~\ref{app:test_time_scaling}): repeated calls produce consistent outputs, so the selector has nothing meaningful to choose between, and additional sampling cannot improve faithfulness. We accordingly do not apply test-time scaling on these tasks.

\textbf{Open-weights image generation models.}
We next ask whether open-weights image generation models can replace Nano-Banana-Pro as the visual tool. We fix Gemini-3.1-Pro as the MLLM and swap in two open-weights alternatives, FLUX.2 [dev]~\citep{flux-2-2025} and Qwen-Image-Edit-2511~\citep{wu2025qwenimagetechnicalreport}, evaluating each across all tasks (Table~\ref{tab:open_gen}). Both models match Nano-Banana-Pro on some tasks, but neither does so consistently across the board. The failure modes differ by model: FLUX.2 struggles with depth estimation, falling below even the no-tools baseline (89.1\% vs.\ 91.7\%), while Qwen-Image-Edit struggles with path tracing (76.0\% vs.\ 91.0\% for Nano-Banana-Pro). For simpler operations such as inpainting, both open-weights models perform competitively, as reflected in strong occluded object counting results (FLUX.2: 7.0\%, Qwen: 6.8\%, vs.\ 7.1\% for Nano-Banana-Pro), suggesting that the gap narrows for tasks with less demanding generative requirements. The broader implication is that current open-weights generators are already useful for local editing-style transformations, but the most spatially precise and globally consistent visualizations still benefit from a stronger image model. Repeating this comparison with an open-weights MLLM, namely Qwen-3.5-27B, yields qualitatively similar results (Appendix~\ref{app:open_gen_qwen}).

\begin{figure*}[t]
    \centering
    \includegraphics[width=0.98\textwidth]{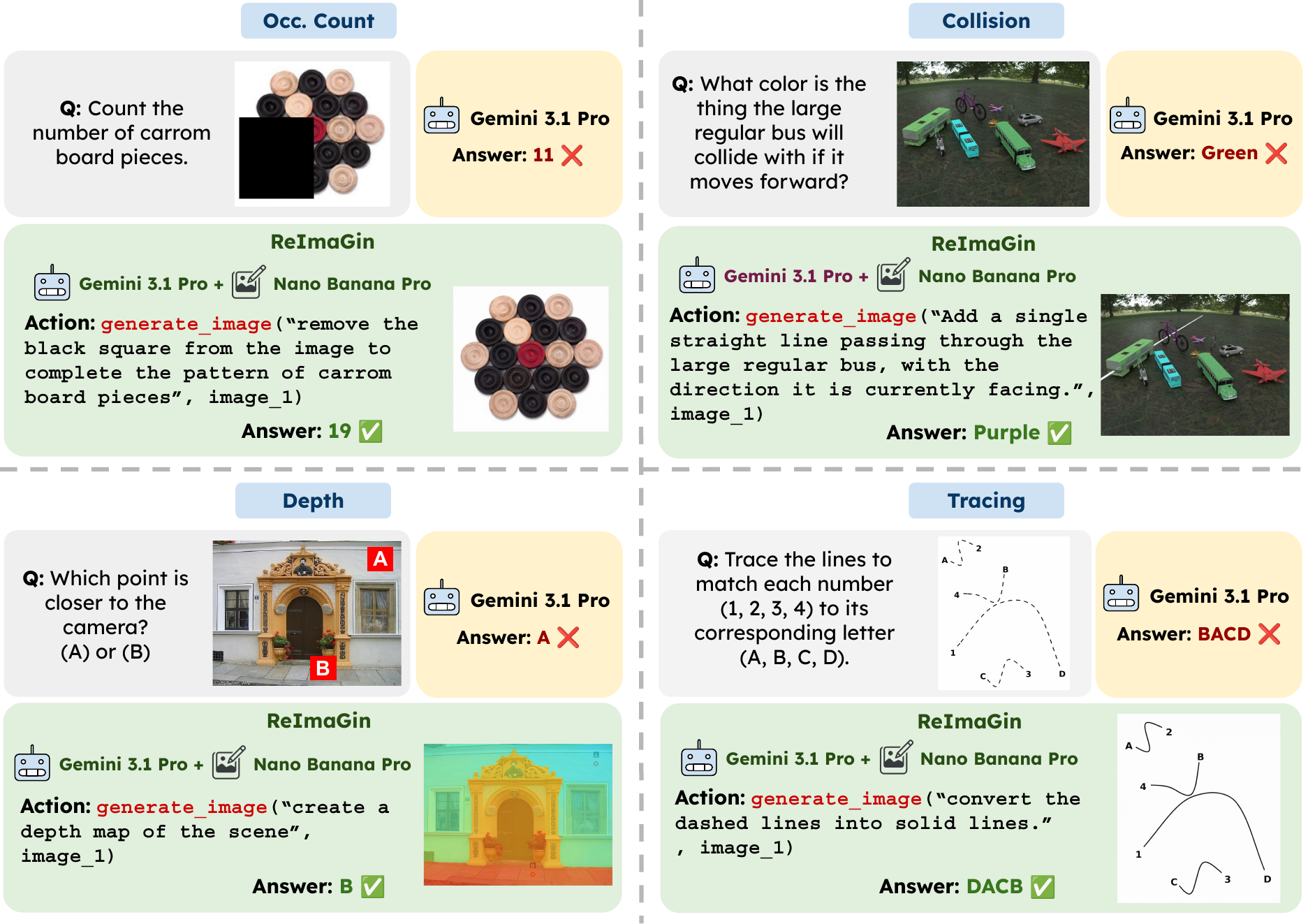}
    \caption{\textbf{Qualitative examples where \ours\ leads to correct reasoning and language-only CoT fails.} For counting, it removes the occluder to reveal the hidden carrom pieces; for collision prediction, it draws the bus trajectory to identify the first object hit; for depth reasoning, it generates a depth map to compare the marked points; and for path tracing, it converts the dashed lines into solid lines to make it easier to trace paths. These examples highlight that \ours{} enables flexible intermediate visual manipulations tailored to the task, and that these manipulations improve reasoning.}
    \label{fig:qualitative}
\end{figure*}

\textbf{Qualitative examples.}
Figure~\ref{fig:qualitative} offers qualitative examples on four tasks
, comparing direct use of Gemini-3.1-Pro vs.\ \ours{}. 
We can see how image generation successfully augments the MLLM's reasoning. Figures~\ref{fig:teaser} and \ref{fig:method} provide further qualitative results.

\textbf{Computational cost.} \ours\ and Sketchpad are matched on the cost and latency of MLLM reasoning (\$$0.04$ per instance), with \ours\ having additional costs for image generation (\$$0.12$ per instance). 
Additional details in Appendix~\ref{app:cost}.

\textbf{Failure modes.} It is worth noting that \ours{} still makes mistakes and struggles with particularly challenging tasks. To quantify how often this happens, we manually audited $120$ generated images spanning all six tasks (roughly $20$ per task), labelling whether each image correctly performs the requested transformation. We find that $75\%$ of generations are faithful, and that faithfulness is strongly associated with task success: the final answer is correct $87\%$ of the time when the generated image is faithful, compared to $43\%$ when it is not. Details of the audit protocol are given in Appendix~\ref{app:faithfulness}. Turning to the nature of these errors, we find that they fall largely into two categories. First, \emph{generation failures}, where the image model produces a visualization that is incorrect, for instance a floorplan that misplaces the relative positions of objects (Figures~\ref{fig:fail_generator_1} and~\ref{fig:fail_generator_2}, Appendix~\ref{app:fail_image_gen}). These errors are the most direct bottleneck and scale with the capability of the underlying generator, as reflected in the gap between Nano-Banana-Pro and open-weights alternatives in Table \ref{tab:open_gen}. Second, \emph{MLLM failures}, where the generated visualization is faithful to the prompt but the MLLM misreads it or fails to reason correctly (Figures~\ref{fig:fail_mllm_1} and~\ref{fig:fail_mllm_2}, Appendix~\ref{app:fail_mllm}). These errors persist even when the generator succeeds, and scale with MLLM capability rather than with generation quality, as shown in Table~\ref{tab:main}.

\begin{table*}[t]
  \centering
  \small
  \resizebox{\linewidth}{!}{
  \begin{tabular}{@{}l cccccc@{}}
  \toprule
  \textbf{Image Gen.\ Model} & \shortstack{\textbf{Puzzle $\uparrow$} \\ {\scriptsize (MIRA)}} & \shortstack{\textbf{Occ.\ Count $\downarrow$} \\ {\scriptsize (CAPTURe)}} & \shortstack{\textbf{Collision $\uparrow$} \\ {\scriptsize (Spatial457)}} & \shortstack{\textbf{Spatial $\uparrow$} \\ {\scriptsize (MMSI)}} & \shortstack{\textbf{Tracing $\uparrow$} \\ {\scriptsize (Ours)}} & \shortstack{\textbf{Depth $\uparrow$} \\ {\scriptsize (BLINK)}} \\
  \midrule
    No Tools              & $37.2_{\pm 2.6}$  & $8.9_{\pm 0.4}$  & $60.3_{\pm 0.9}$  &  $49.0_{\pm 0.6}$  &  $70.7_{\pm 1.2}$  & $91.7_{\pm 0.3}$ \\
    Nano-Banana-Pro       & $42.3_{\pm 2.2}$  & $7.1_{\pm 0.6}$  & $\textbf{69.7}_{\pm 1.2}$  & $\textbf{59.0}_{\pm 0.6}$  &  $\textbf{91.0}_{\pm 1.7}$  & $\textbf{94.6}_{\pm 1.5}$ \\
    FLUX.2 [dev]          & $\textbf{53.8}_{\pm 2.2}$  & $7.0_{\pm 0.3}$  & $62.0_{\pm 1.0}$  & $54.0_{\pm 0.9}$  & $87.0_{\pm 1.0}$  & $89.1_{\pm 0.4}$ \\
    Qwen-Image-Edit-2511  & $38.5_{\pm 2.2}$  & $\textbf{6.8}_{\pm 0.2}$  & $68.0_{\pm 0.8}$  & $54.5_{\pm 1.8}$  & $76.0_{\pm 4.0}$  & $94.0_{\pm 1.2}$ \\
  \bottomrule
  \end{tabular}
  }
  \caption{
  \textbf{Effect of the image generation model on \ours\ performance, with Gemini-3.1-Pro as the MLLM.}
  Nano-Banana-Pro is compared against two open-weights alternatives.
  Open-weights models help on some tasks but lag behind overall, with different failure modes.
  Open generators are useful for simpler editing-style transformations but less reliable on spatially precise tasks.
  }
  \label{tab:open_gen}
\end{table*}

\subsection{Automated Discovery of Visual Reasoning Strategies}
\input{tables/results_discovered}
\input{tables/results_strategy_transfer}
Reasoning with generated images requires a strategy for deciding when a visualization is needed, what visualization to generate, and how to use it afterward.
We therefore explore whether such strategies can also be discovered automatically, following our approach presented in Section~\ref{sec:discovery}; results appear in Table~\ref{tab:genimage_ablation}.
We evaluate discovery in two regimes.
The first pairs the strong Gemini-3.1-Pro MLLM~\citep{gemini2025gemini3}, used as both proposal model and task-solving agent, with the strong Nano-Banana-Pro generator~\citep{2025nanobananapro}.
The second pairs the weaker open-weights Qwen-3.5-27B MLLM~\citep{qwen35blog}, again used for both proposal and task solving, with the flash version of the image generator, Nano-Banana-2 (Gemini-3.1-Flash-Image;~\citet{google2026gemini31flashimage}).
The discovered strategy outperforms No Tools on all five tasks with Gemini and four of five with Qwen (tying on depth).
With the stronger pairing, it also consistently outperforms the No Strategy setup, where the MLLM is given access to the generative image tool but no specific strategy.
Together, these results show that the discovery procedure can produce useful task strategies across varying reasoning and image-generation model capabilities.

\begin{figure*}[t]
    \centering
    \includegraphics[width=0.96\textwidth]{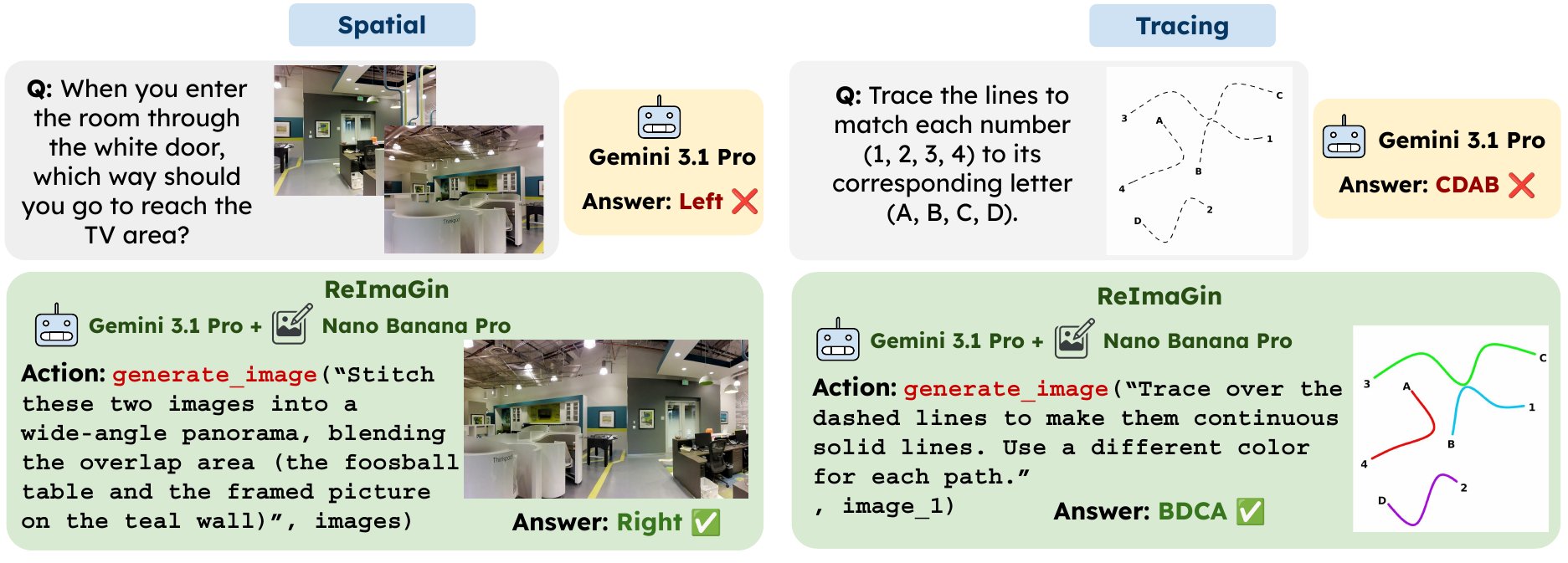}
    \caption{
        \textbf{Automatically discovered visual reasoning strategies.}
        Discovered strategies are frequently similar to handcrafted ones (generate a panorama, solid lines), with some differences (color the lines).
        Appendix~\ref{app:discovered_strategy_failures} contains more qualitative examples of success and failure cases of discovered strategies.
    }
    \label{fig:discovered_strategy_qualitative}
    \vspace{-5pt}
\end{figure*}

As illustrated in Figure~\ref{fig:discovered_strategy_qualitative}, discovered strategies often closely match the handcrafted ones.
For instance, the optimized depth prompt learns to generate a depth map, and the collision prompt learns to draw an arrow from the front of the object that will collide with another.
These small differences in wording result in only small drops relative to the handcrafted baselines.
In cases where the difference between the discovered and handcrafted strategies is larger, the gap widens: for path tracing, our handcrafted strategy simply makes dashed lines solid, whereas the optimized strategy additionally colors them.
Because Nano-Banana-Pro rarely fails to make lines solid but more frequently fails to color them correctly, this richer strategy slightly degrades performance.
Similarly, the composite strategy we use for MMSI spatial reasoning is not recovered under the current search setting, although a useful strategy that requests a panorama-like view of the room is discovered.
The independently Qwen-discovered prompts exhibit the same qualitative pattern (Appendix~\ref{app:strategy_prompts}).
Appendix~\ref{app:strategy_prompts} provides the discovery meta prompt and the discovered strategies.

\textbf{Cross-MLLM transfer.}
We further test whether strategy discovery is tied to the MLLM used during optimization, or whether a discovered strategy can in fact be used by a different MLLM with different reasoning capabilities.
To test this, we use the Gemini-3.1-Pro-discovered prompts unchanged with Qwen-3.5-27B, without rediscovery, as shown in Table~\ref{tab:strategy_transfer}.
The transferred strategies outperform No Tools on all five tasks, and Visual Sketchpad and No Strategy on collision, spatial reasoning (MMSI), and path tracing.
Because the prompts are applied unchanged, these gains show that their utility generalizes beyond the MLLM used during discovery.
However, as seen by the degraded performance on Occlusion counting and the lack of improvement in Depth, a visual reasoning strategy optimized for an MLLM with strong perceptual and reasoning capabilities can sometimes fall short of bringing the same benefits when paired with a weaker model.

%% file: tables/results_discovered.tex
\begin{table*}[t]
\centering
\footnotesize
\resizebox{\textwidth}{!}{%
\begin{tabular}{@{}lll ccccc@{}}
\toprule
\textbf{MLLM} & \textbf{Image generator} & \textbf{Approach} & \shortstack{\textbf{Occ.\ Count $\downarrow$} \\ {\scriptsize (CAPTURe)}} & \shortstack{\textbf{Collision $\uparrow$} \\ {\scriptsize (Spatial457)}} & \shortstack{\textbf{Spatial $\uparrow$} \\ {\scriptsize (MMSI)}} & \shortstack{\textbf{Tracing $\uparrow$} \\ {\scriptsize (Ours)}} & \shortstack{\textbf{Depth $\uparrow$} \\ {\scriptsize (BLINK)}} \\
\midrule
\multirow{3}{*}{Gemini-3.1-Pro}
  & \multirow{3}{*}{Nano-Banana-Pro} & \gr{Handcrafted Strategy} & \gr{$7.1_{\pm 0.6}$} & \gr{$69.7_{\pm 1.2}$} & \gr{$59.0_{\pm 0.6}$} &  \gr{$91.0_{\pm 1.7}$} & \gr{$94.6_{\pm 1.5}$} \\
  \cmidrule(l){3-8}
  & & No Strategy        & $12.1_{\pm 0.3}$ & $60.0_{\pm 0.0}$ & $49.3_{\pm 0.3}$ & $72.8_{\pm 0.3}$ & $90.3_{\pm 0.0}$ \\
  & & Automatic Strategy & $\textbf{8.0}_{\pm 0.2}$ & $\textbf{64.0}_{\pm 0.6}$ & $\textbf{56.7}_{\pm 0.7}$ & $\textbf{87.3}_{\pm 0.9}$ & $\textbf{94.3}_{\pm 0.5}$ \\
\midrule
\multirow{2}{*}{Qwen-3.5-27B}
  & \multirow{2}{*}{Nano-Banana-2} & No Strategy & $9.6_{\pm 0.6}$ & $48.0_{\pm 2.1}$ & $42.4_{\pm 1.0}$ & $42.1_{\pm 4.4}$ & $\textbf{86.0}_{\pm 1.5}$ \\
  &  & Automatic Strategy & $\textbf{9.5}_{\pm 0.3}$ & $\textbf{54.9}_{\pm 1.7}$ & $\textbf{44.8}_{\pm 1.6}$ & $\textbf{44.8}_{\pm 4.0}$ & $85.5_{\pm 0.9}$ \\
\bottomrule
\end{tabular}
}
\caption{
\textbf{\ours{} with automatic visual reasoning strategy discovery.}
``Handcrafted Strategy'' uses a human-designed visual reasoning policy, ``No Strategy'' employs image generation without task-specific guidance, and ``Automatic Strategy'' uses a policy discovered without human guidance.
For each automatic-discovery row, the listed MLLM serves as both proposal model and task-solving agent.
With Gemini, Automatic Strategy improves over No Strategy on all five tasks and recovers a substantial fraction of the handcrafted gains.
For Qwen, Automatic Strategy improves over No Strategy in most tasks, effectively tying on two benchmarks, namely Occ. Counting and Depth.
}
\label{tab:genimage_ablation}
\end{table*}

%% file: tables/results_strategy_transfer.tex
\begin{table*}[t]
\centering
\footnotesize
\setlength{\tabcolsep}{7pt}
\begin{tabular}{@{}lccccc@{}}
\toprule
\textbf{Qwen-3.5-27B setting} & \textbf{Occ.\ Count $\downarrow$} & \textbf{Collision $\uparrow$} & \textbf{Spatial $\uparrow$} & \textbf{Tracing $\uparrow$} & \textbf{Depth $\uparrow$} \\
\midrule
No Tools                          &  $12.1_{\pm 0.3}$              &  $\underline{51.7}_{\pm 1.5}$ &  $42.0_{\pm 1.5}$              & $35.7_{\pm 1.2}$              &  $85.5_{\pm 2.0}$ \\
Visual Sketchpad                  & $\underline{8.7}_{\pm 0.0}$  &  $48.3_{\pm 1.8}$              &  $\underline{43.0}_{\pm 0.6}$ &  $28.0_{\pm 4.5}$              &  $\mathbf{89.5}_{\pm 1.6}$ \\
No Strategy                       & $\mathbf{8.3}_{\pm 0.7}$     & $48.2_{\pm 2.2}$              & $42.0_{\pm 1.3}$                     & $\underline{42.3}_{\pm 4.9}$ & $85.5_{\pm 1.3}$ \\
Gemini-discovered Strategy        & $10.9_{\pm 0.3}$              & $\mathbf{54.4}_{\pm 1.2}$     & $\mathbf{47.8}_{\pm 1.7}$            & $\mathbf{55.9}_{\pm 3.4}$     & $\underline{85.8}_{\pm 0.7}$ \\
\bottomrule
\end{tabular}
\caption{
\textbf{Cross-MLLM transfer of automatically discovered strategies.}
We apply the task-specific prompts discovered with Gemini-3.1-Pro directly to Qwen-3.5-27B, paired with Nano-Banana-Pro as the image generator.
Visual Sketchpad instead uses its fixed specialist tools, and No Tools has no tool access.
In most tasks, gains over text-only reasoning and other settings generalize to a new model.
}
\label{tab:strategy_transfer}
\end{table*}

%% file: files/conclusion.tex
\section{Conclusion}
We introduced \ours, a multimodal reasoning framework that uses image generation as a flexible mechanism for visual reasoning.
Rather than restricting reasoning to text or relying on a fixed set of specialist visual tools, our approach allows the model to \emph{generate} open-ended visual transformations to support subsequent reasoning.
Across a broad evaluation covering six visual reasoning tasks and multiple model backbones, \ours\ consistently improves over both text-only reasoning and a strong specialist-tool baseline.
This suggests that the benefit is not tied to a single task, transformation, or model.

Our results suggest two broader conclusions.
First, image generation can act as a general visual imagination mechanism within multimodal reasoning, reducing the need to engineer a separate tool for each new transformation.
Second, the effectiveness of this approach depends not only on the underlying models but also on the reasoning strategy used to decide what to visualize and how to use the result.
The gains from automatic strategy discovery show that the models can leverage their understanding of visual transformations to discover relevant strategies without human-authored task-specific strategies or reasoning.

An important next step is improving the reliability of generative visual reasoning.
Our results suggest that this can come from stronger image generation models, better selection and verification of intermediate visualizations, and more scalable methods for discovering effective reasoning strategies.
Overall, we view reasoning with image generation as a promising step toward multimodal systems that can not only describe visual scenes, but also actively generate and transform them while solving complex problems.

\section*{Acknowledgments}

The research was partially funded by a LOEWE Start-Professur (LOEWE/4b//519/05.01.002(0006)/94), a Spitzen-Professur
(LOEWE/4a//519/05.00.002-(0010)/93), and an Alexander von Humboldt Professorship in Multimodal Reliable AI, sponsored by the German Federal Ministry of Research, Technology and Space (BMFTR) and has benefited from the Excellence Cluster ``Reasonable AI'' by the German Research Foundation (Deutsche Forschungsgemeinschaft - DFG) under Germany’s Excellence Strategy – EXC-3057. We gratefully acknowledge support from the hessian.AI Service Center (funded by the BMFTR, grant no. 16IS22091) and the hessian.AI Innovation Lab (funded by the Hessian Ministry for Digital Strategy and Innovation,
grant no. S-DIW04/0013/003). We also gratefully acknowledge the Gemini Academic Program for providing Gemini API credits. We also thank Hritik Bansal and Ashima Suvarna for helpful feedback on the manuscript.

%% file: files/appendix.tex
\section*{LLM Use Statement}
The authors used large language model (LLM) tools to assist with language refinement and polishing of the manuscript text, as well as for LaTeX scripting and formatting.

\section{Extended Related Work}
\label{app:related_work}

\input{files/related_work_extended}

\section{Benchmark Details}
\label{app:benchmarks}

\textbf{Depth Perception.}
We use the validation split of BLINK~\citep{fu2024blink}, containing 124 instances, following the evaluation protocol of Visual Sketchpad~\citep{hu2024visual}.

\textbf{Puzzle Completion.}
We use the test split of MIRA~\citep{zhou2025visualizing}, containing 26 instances.

\textbf{Occlusion Counting.}
We randomly sample 100 instances from the CAPTURe benchmark~\citep{pothiraj2025capture}.

\textbf{Collision Prediction.}
We randomly sample 100 instances from Spatial457~\citep{wang2025spatial457}.

\textbf{Multi-View Spatial Reasoning.}
We randomly sample 50 instances each from the Obj-Obj and Obj-Reg splits of MMSI-Bench~\citep{yang2025mmsi}, for a total of 100 instances.

\textbf{Path Tracing.}
We programmatically generate 100 instances. To increase difficulty, the connecting lines are rendered as dashed rather than solid, and at least one pair of lines is designed to pass close to each other without intersecting.

\section{Method Overview}
\label{app:overview}

\begin{figure*}[h]
    \centering
    \includegraphics[width=0.8\textwidth]{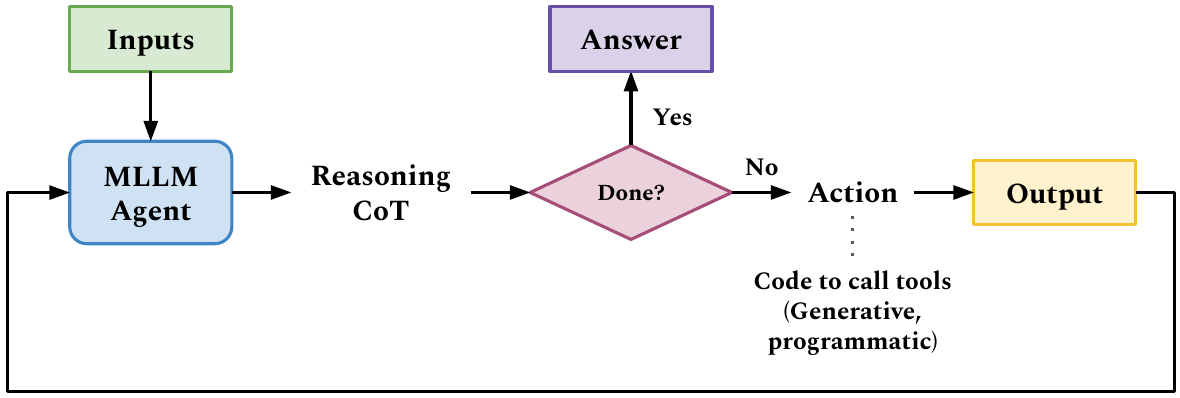}
    \caption{\textbf{Overview of \ours.} The agent alternates between reasoning and tool actions, feeds outputs back into context, and stops when ready to answer.}
    \label{fig:overview}
\end{figure*}

\section{Additional Test-Time Scaling Results}
For MMSI, we apply the same image-generation test-time scaling used by the handcrafted strategy to both Automatic Strategy and No Strategy.
Relative to single-candidate evaluation, this changes Automatic Strategy from $55.7_{\pm 1.2}$ to $56.7_{\pm 0.7}$ with Gemini and from $44.4_{\pm 0.5}$ to $44.8_{\pm 1.6}$ with Qwen/Flash.
For Gemini No Strategy, the aggregate-only result changes from $51.2$ to $49.3_{\pm 0.3}$.

\section{Illustrative Example}
\label{app:worked_ex}

We illustrate one full reasoning trajectory on the \textsc{Spatial457} collision-prediction task (Figure~\ref{fig:qualitative} top right). The input is an overhead scene image containing several coloured toy vehicles and the query \emph{``What shape is the brown object that the large brown aeroplane hits if it moves forward?''}

\begin{enumerate}
    \item \textbf{Turn 1 --- Reasoning:} The agent identifies the large brown aeroplane in the scene and notes that its forward trajectory is ambiguous because multiple objects lie in roughly the same direction. It decides that a generated overlay showing a single forward vector from the nose of the aeroplane would resolve the ambiguity.

    \item \textbf{Turn 1 --- Action:} The agent calls
    \begin{verbatim}
generate_image(
  prompt="Draw one forward vector from the nose of the
          large brown aeroplane and circle the first
          object it hits. Lightly label brown objects only.",
  images=[scene_image]
)
    \end{verbatim}
    The tool returns an annotated image with a forward arrow and a circled target.

    \item \textbf{Turn 2 --- Context:} The annotated image is appended to the context as a new user message alongside the original scene.

    \item \textbf{Turn 2 --- Reasoning:} Examining the overlay, the agent observes that the arrow terminates at a rectangular brown object. It verifies the label against the original scene image to confirm the object's identity, then emits its final answer.

    \item \textbf{Turn 2 --- Termination:} The agent outputs \emph{``The first object hit is a brown minivan. \textsc{Terminate}''}, and the answer \emph{Minivan} is extracted.
\end{enumerate}

Without the intermediate visualization, a no-tools baseline must reason about 3D trajectories purely from the original image, frequently confusing the aeroplane's heading with nearby objects and predicting the wrong collision target. The generated overlay reduces the problem to a simple visual lookup, which the MLLM handles reliably.

\section{Test-Time Scaling via Repeated Sampling}
\label{app:test_time_scaling}

\begin{figure}[h]
    \centering
    \includegraphics[width=0.7\linewidth]{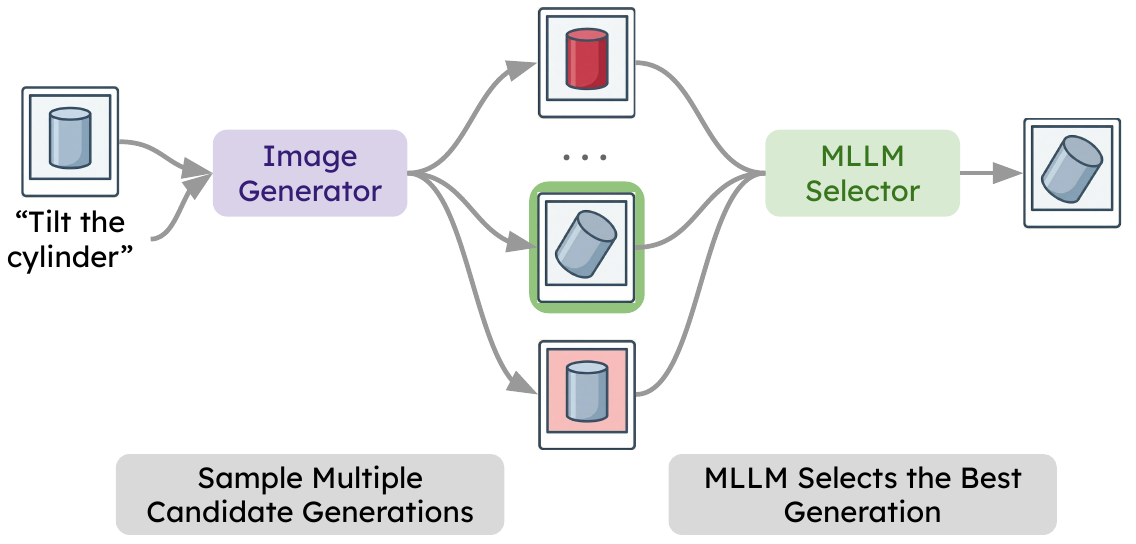}
    \caption{\textbf{Test-time scaling via repeated sampling.} Given a prompt and optional reference images, the image generator produces $N$ candidate outputs independently. An MLLM selector then evaluates all candidates jointly and returns the one that most faithfully follows the prompt. The selector is simply an MLLM, in our case Gemini-3.1-Pro, prompted to select the best generation. See prompt in Appendix~\ref{app:selector_prompt}.}
    \label{fig:test_time_scaling}
\end{figure}

\begin{figure*}[h]
    \centering
    \includegraphics[width=0.56\textwidth]{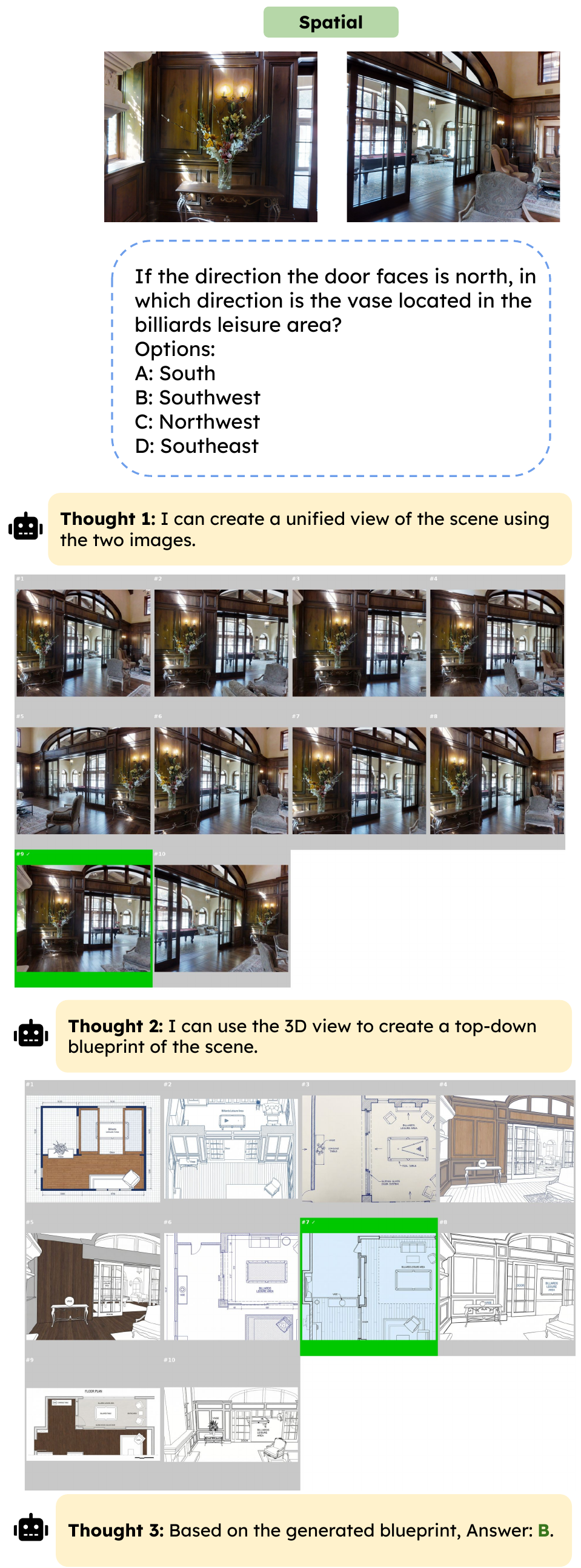}
    \caption{\textbf{Test time scaling: the image generation model is called 10 times with the same prompt, outputting varied images.} A selection model (Gemini 3.1 Pro) is used to choose the image that is most faithful to the prompt, here indicated in green.}
    \label{fig:test_time_scaling_result}
\end{figure*}

\begin{figure*}[h]
    \centering
    \includegraphics[width=0.7\linewidth]{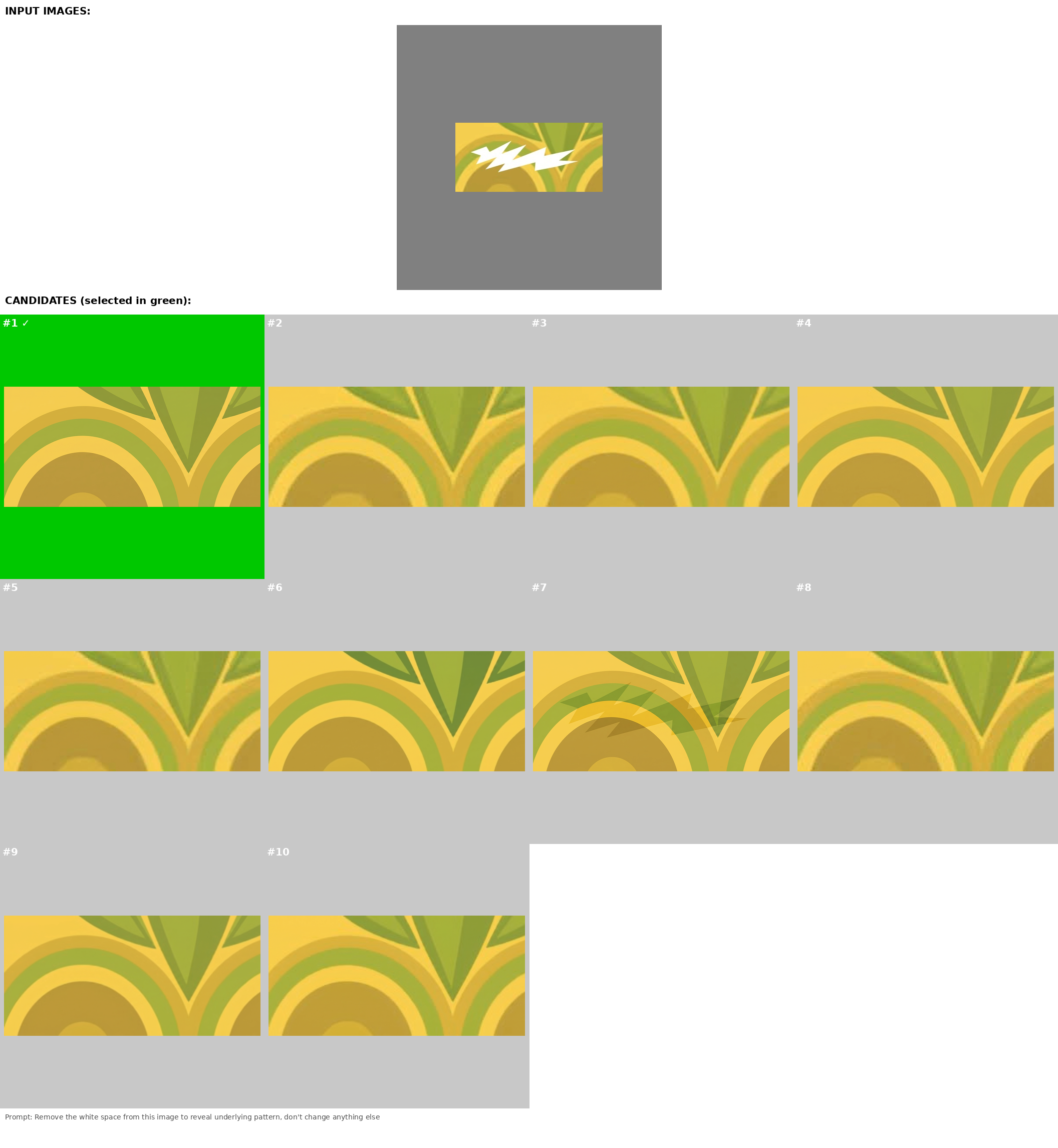}
    \caption{\textbf{For the puzzle task, repeated samples from the image generator are consistent with each other, leaving little room for test-time scaling to improve performance.} We therefore do not apply test-time scaling to this task.}
    \label{fig:test_time_scaling_mira}
\end{figure*}

\begin{figure*}[h]
    \centering
    \includegraphics[width=0.56\textwidth]{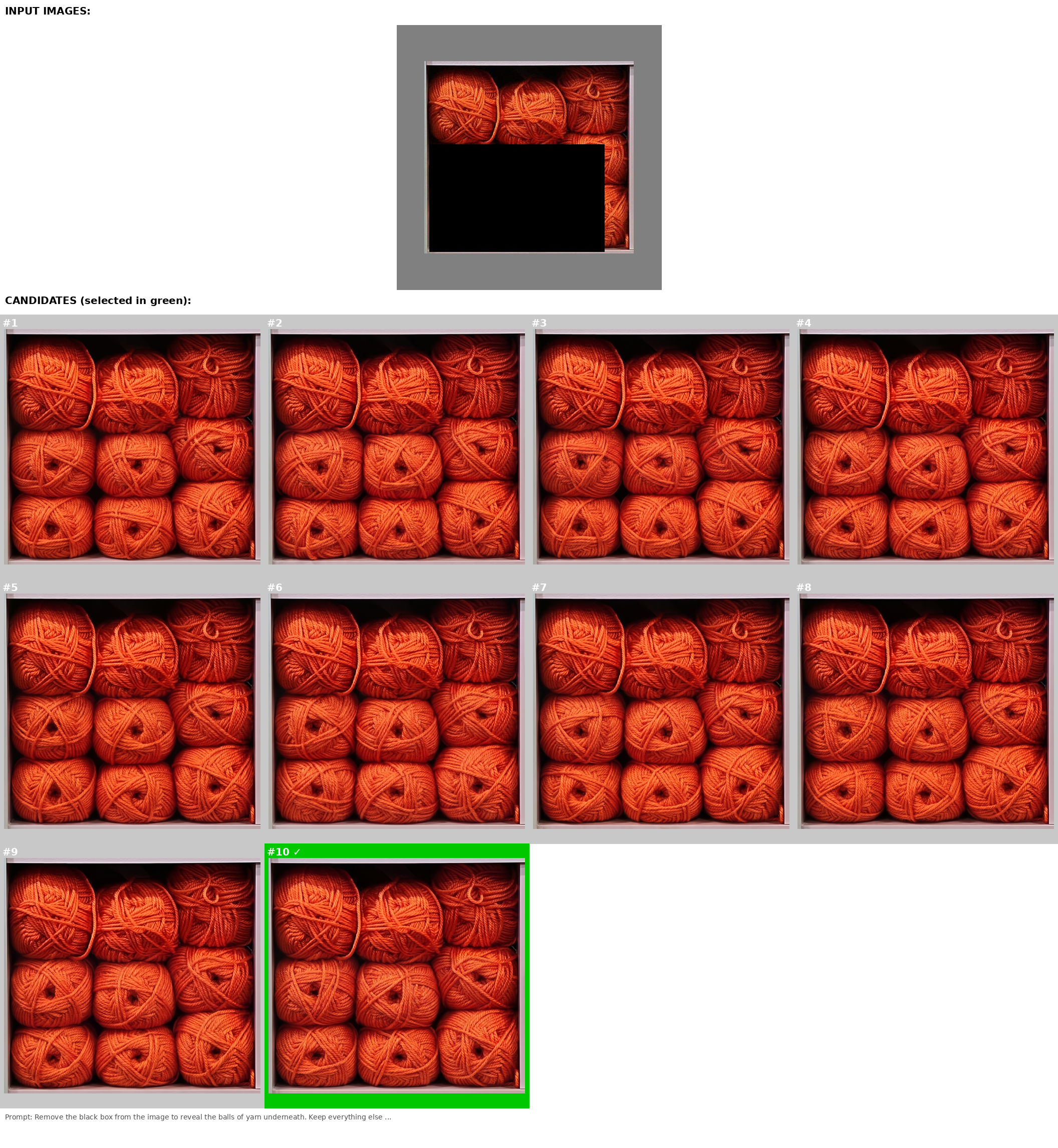}
    \caption{\textbf{For the occlusion task, repeated samples from the image generator are consistent with each other, leaving little room for test-time scaling to improve performance.} We therefore do not apply test-time scaling to this task.}
    \label{fig:test_time_scaling_capture_occlusion}
\end{figure*}

Figure~\ref{fig:test_time_scaling} shows how repeated sampling is used to improve the accuracy of generated images. Figure~\ref{fig:test_time_scaling_result} further illustrates the stochasticity in repeated calls of the image generator in case of the spatial reasoning task. On the other hand, other tasks do not show such variance among samples, as shown in Figures~\ref{fig:test_time_scaling_mira} and~\ref{fig:test_time_scaling_capture_occlusion}, leaving little room for test-time scaling to improve the faithfulness of the generator output. Hence, we only apply test-time scaling to the spatial reasoning task.

\section{Test-Time Scaling for Baselines}
\label{app:baseline_voting}
On MMSI, \ours\ spends additional compute on image generation: each image generation call samples $10$ candidate images and a selector keeps one, while a single reasoning trace is sampled. To test whether extra compute alone closes the gap, we give the two baselines additional test-time compute via majority voting over $20$ sampled answers (Table~\ref{tab:baseline_voting}). Each voted baseline uses $3$ disjoint groups of $20$ runs (mean $\pm$ standard error over groups). Voting narrows the gap for Gemini-3.1-Pro and closes it for Qwen-3.5-27B, where voted Sketchpad matches \ours\ ($48.7\%$ vs.\ $48.0\%$), but not for GPT-5, where voting lowers Sketchpad's accuracy and the gap widens.

\begin{table}[h]
  \centering
  \small
  \begin{tabular}{@{}ll ccc@{}}
  \toprule
  \textbf{Approach} & \textbf{Test-time compute} & \textbf{Gemini-3.1-Pro} & \textbf{GPT-5} & \textbf{Qwen-3.5-27B} \\
  \midrule
  No Tools  & single sample          & $49.0_{\pm 0.6}$ & $36.0_{\pm 2.0}$ & $42.0_{\pm 1.5}$ \\
  No Tools  & majority vote ($20$)   & $53.7_{\pm 1.5}$ & $38.0_{\pm 0.6}$ & $44.7_{\pm 0.7}$ \\
  Sketchpad & single sample          & $55.7_{\pm 3.5}$ & $40.3_{\pm 2.4}$ & $43.0_{\pm 0.6}$ \\
  Sketchpad & majority vote ($20$)   & $57.3_{\pm 1.7}$ & $36.7_{\pm 0.9}$ & $\textbf{48.7}_{\pm 2.3}$ \\
  \rowcolor{blue!10} \ours & best-of-$10$ images & $\textbf{59.0}_{\pm 0.6}$ & $\textbf{43.7}_{\pm 3.2}$ & $48.0_{\pm 4.0}$ \\
  \bottomrule
  \end{tabular}
  \caption{\textbf{Test-time scaling for baselines on MMSI (accuracy, \%).} Majority voting over $20$ sampled answers narrows the gap to \ours\ for Gemini-3.1-Pro, closes it for Qwen-3.5-27B, and does not close it for GPT-5.}
  \label{tab:baseline_voting}
\end{table}

\clearpage
\section{Computational Cost}
\label{app:cost}

We report the per-instance inference cost of \ours\ and the two baselines in Table~\ref{tab:cost}, measured with Gemini-3.1-Pro as the MLLM and Nano-Banana-Pro as the visual generative tool, averaged over all task instances.

\ours\ and Sketchpad are matched on every quantity that reflects reasoning effort: MLLM input tokens ($19{,}700$ vs.\ $21{,}000$), output tokens ($350$ vs.\ $400$), MLLM calls ($1.79$ vs.\ $1.83$), and intermediate images produced ($1.2$ vs.\ $1.3$), which yield an identical MLLM cost of \$$0.04$ per instance for both methods.
Both methods therefore spend essentially the same effort on reasoning, and neither obtains its accuracy by simply issuing more or longer MLLM calls.
The two differ only in the cost and latency of image generation: Sketchpad's intermediate images are produced by specialist vision models and incur no generation-token cost, whereas \ours\ synthesizes them with a frontier image model, adding $1{,}300$ generation tokens and \$$0.12$ per instance and raising median wall-clock time from $52$\,s to $76$\,s.
This is the price of replacing a fixed set of specialist tools with a single general-purpose generative one, and we expect it to fall as generative models become cheaper and faster.

This accounting covers inference only; the one-time cost of automated strategy discovery is reported separately in Appendix~\ref{app:strategy_prompts}.

\input{tables/results_cost}

\clearpage
\section{Open-Weights Generators with an Open-Weights MLLM}
\label{app:open_gen_qwen}

Table~\ref{tab:open_gen} compares image generation models with Gemini-3.1-Pro as the MLLM. To check whether those conclusions depend on a proprietary MLLM, we repeat the comparison with Qwen-3.5-27B, so that the FLUX.2 [dev] and Qwen-Image-Edit-2511 rows correspond to a fully open-weights pipeline.

The results in Table~\ref{tab:open_gen_qwen} closely track the Gemini setting. Nano-Banana-Pro remains the strongest generator on five of six tasks. Both open-weights alternatives improve substantially over the No Tools baseline on occlusion counting ($6.1$ and $8.8$ versus $12.1$ sMAPE) and path tracing ($66.0$ and $48.0$ versus $35.7$), and FLUX.2 [dev] also on puzzle completion ($44.2$ versus $34.6$, close to $46.2$ for Nano-Banana-Pro); FLUX.2 [dev] achieves the best occlusion counting result overall ($6.1$ sMAPE). The exceptions are collision prediction, spatial reasoning, and depth, where both open generators perform at or below No Tools ($45.5$ and $50.0$ versus $51.7$ on collision; $38.0$ and $42.3$ versus $42.0$ on spatial; $85.5$ and $80.3$ versus $85.5$ on depth). This matches the pattern observed with Gemini-3.1-Pro: open-weights generators handle editing-style transformations well but remain unreliable on the spatially precise operations that these tasks require. That the same ordering holds under a weaker, open-weights MLLM suggests the limitation lies with the generators rather than with the reasoning model.

\input{tables/results_open_gen_qwen}

\clearpage
\section{Ablating the Generative Tool}
\label{app:no_generate}

The baselines in Table~\ref{tab:main} differ from \ours\ in more than tool access: No Tools and Visual Sketchpad use their own prompts and their own agent scaffolds, so a comparison against them cannot by itself attribute the observed gains to image generation specifically. To rule out these confounds, we ablate the generative tool within \ours\ itself while holding everything else fixed. Concretely, we replace \texttt{generate\_image} with a no-op that simply returns the input image it received, unchanged. The tool therefore remains available to the agent and is still invoked as usual, so the prompts, the tool definitions, the Python execution environment, and the programmatic image utilities are all identical to \ours; the only difference is that calling the tool no longer produces a new visualization. This isolates the generative capability as the single variable that changes.

As shown in Table~\ref{tab:no_generate}, performance drops on all six tasks, with the largest decreases on path tracing ($91.0$ to $77.5$) and puzzle completion ($42.3$ to $36.5$). The ablated variant still outperforms the No Tools baseline in Table~\ref{tab:main} on several tasks, which is expected given that it retains code execution and programmatic image manipulation, but it does not recover the full performance of \ours. Since the prompts and scaffold are identical across the two rows, the remaining difference is attributable to the generative tool alone, and not to prompt or scaffold differences between \ours\ and the baselines.

\input{tables/results_no_generate}

\clearpage
\section{Faithfulness of the Generated Images}
\label{app:faithfulness}

\ours\ depends on the generative model producing an image that actually performs the transformation the agent requested. To measure how often this holds, we manually audited $120$ generated images sampled across all six tasks, roughly $20$ per task. Each image was labelled as \emph{faithful} if it correctly carries out the requested transformation, for example if the occluder is genuinely removed, the trajectory line follows the stated heading, or the top-down map preserves the relative positions of objects, and \emph{unfaithful} otherwise. We then paired each label with whether the agent ultimately answered the question correctly. For MMSI, where \ours\ applies test-time scaling, the audited image is the one chosen by the selector rather than an arbitrary sample, so the reported faithfulness reflects the images the MLLM actually reasons over.

Of the audited generations, $75\%$ are faithful. Conditioning on this label separates task outcomes sharply: the agent answers correctly in $87\%$ of instances with a faithful image, compared to $43\%$ of instances with an unfaithful one. Unfaithful generations therefore account for a substantial share of the remaining errors, consistent with the generation failures illustrated in Appendix~\ref{app:fail_image_gen}.

This relationship is correlational, since instances that are harder for the generator may also be harder for the MLLM to reason about. The test-time scaling experiment in Figure~\ref{fig:tts_bar_chart} provides complementary evidence: sampling $N=10$ images and selecting the most faithful one improves spatial reasoning accuracy from $53.3\%$ to $59.0\%$ on Gemini-3.1-Pro and from $41.7\%$ to $48.0\%$ on Qwen-3.5-27B, while leaving the textual reasoning process untouched. Because the additional compute is spent only on sampling images, the resulting gain is attributable to improved faithfulness rather than to more reasoning.

\clearpage
\section{Examples of Failure Cases}\label{sec:failure}

\subsection{Failures of Image Generator}
\label{app:fail_image_gen}

\begin{figure*}[h]
    \centering
    \includegraphics[width=0.8\textwidth]{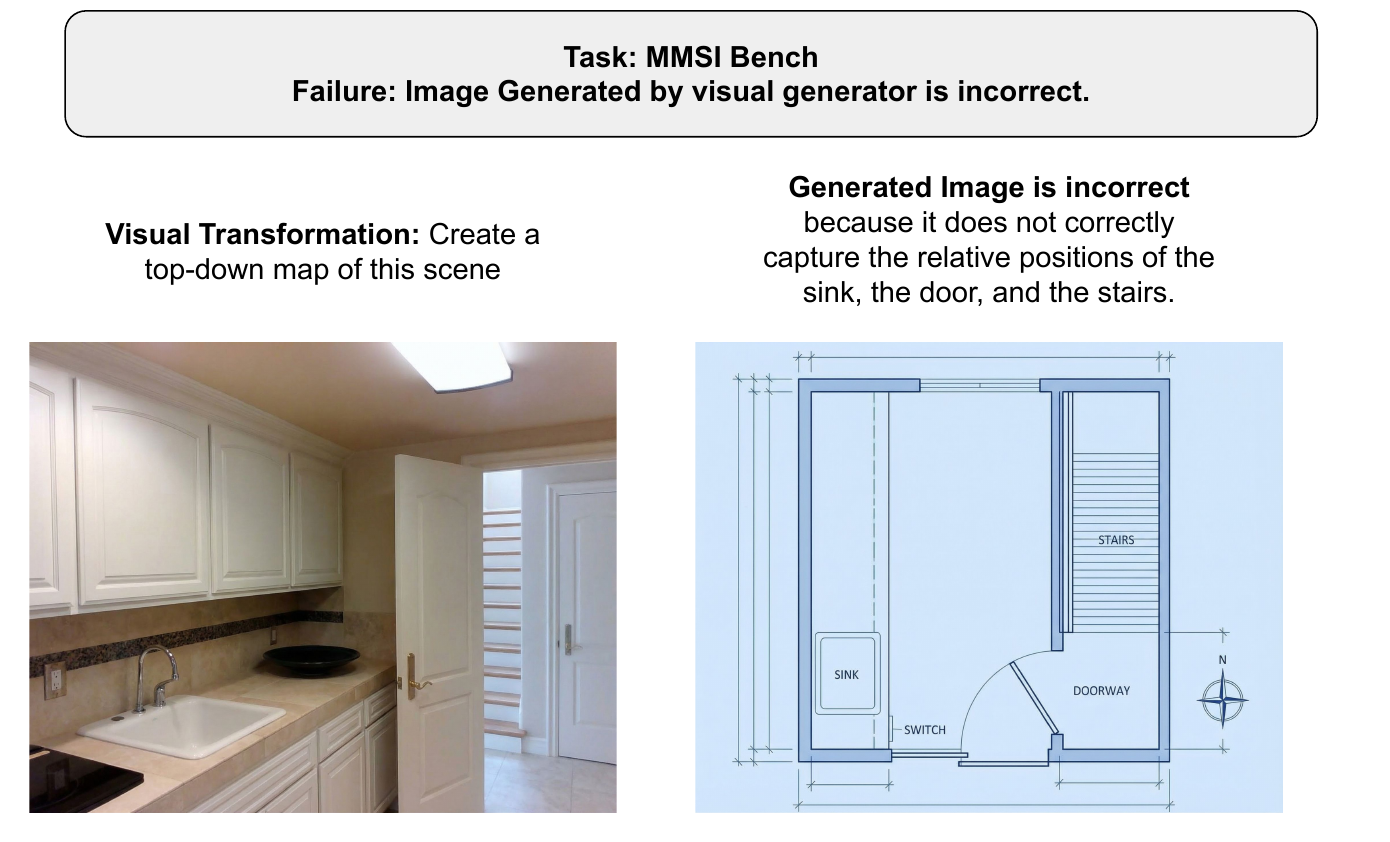}
    \caption{\textbf{Example where the generative model fails to generate the correct visualization.} The task for the generative model is to create a top-down map of the scene. However, the output does not capture the relative positions of different objects correctly.}
    \label{fig:fail_generator_1}
\end{figure*}

\begin{figure*}[h]
    \centering
    \includegraphics[width=0.8\textwidth]{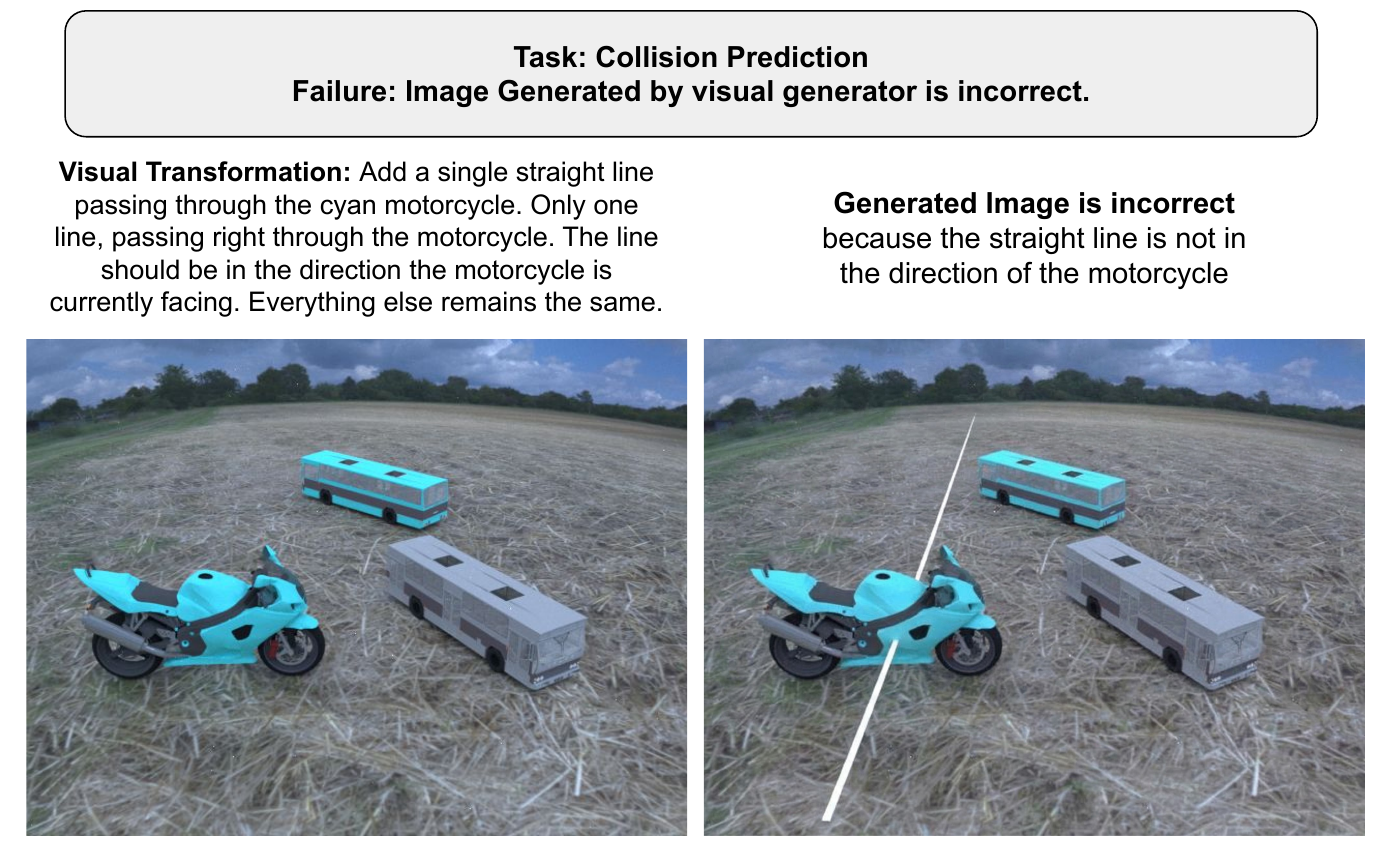}
    \caption{\textbf{Example where the generative model fails to generate the correct visualization.} The task for the generative model is to draw a line through the cyan motorcycle in the direction the motorcycle is facing. However, the generated image shows a line perpendicular to the direction of the motorcycle.}
    \label{fig:fail_generator_2}
\end{figure*}

Here, we provide some qualitative examples where the image generator fails to generate the correct visualization. In Figure~\ref{fig:fail_generator_1}, the task is to create a top-down map of the scene. Here, the generated image fails to capture the relative positions of the sink, doorway, and the stairs in the input image. In Figure~\ref{fig:fail_generator_2}, task is to draw a line going through the cyan motorcycle in the direction faced by the motorcycle. Here, the line passes through the motorcycle, but its direction is perpendicular to the direction of the motorcycle.

\subsection{Failures of the MLLM}
\label{app:fail_mllm}

\begin{figure*}[h]
    \centering
    \includegraphics[width=0.8\textwidth]{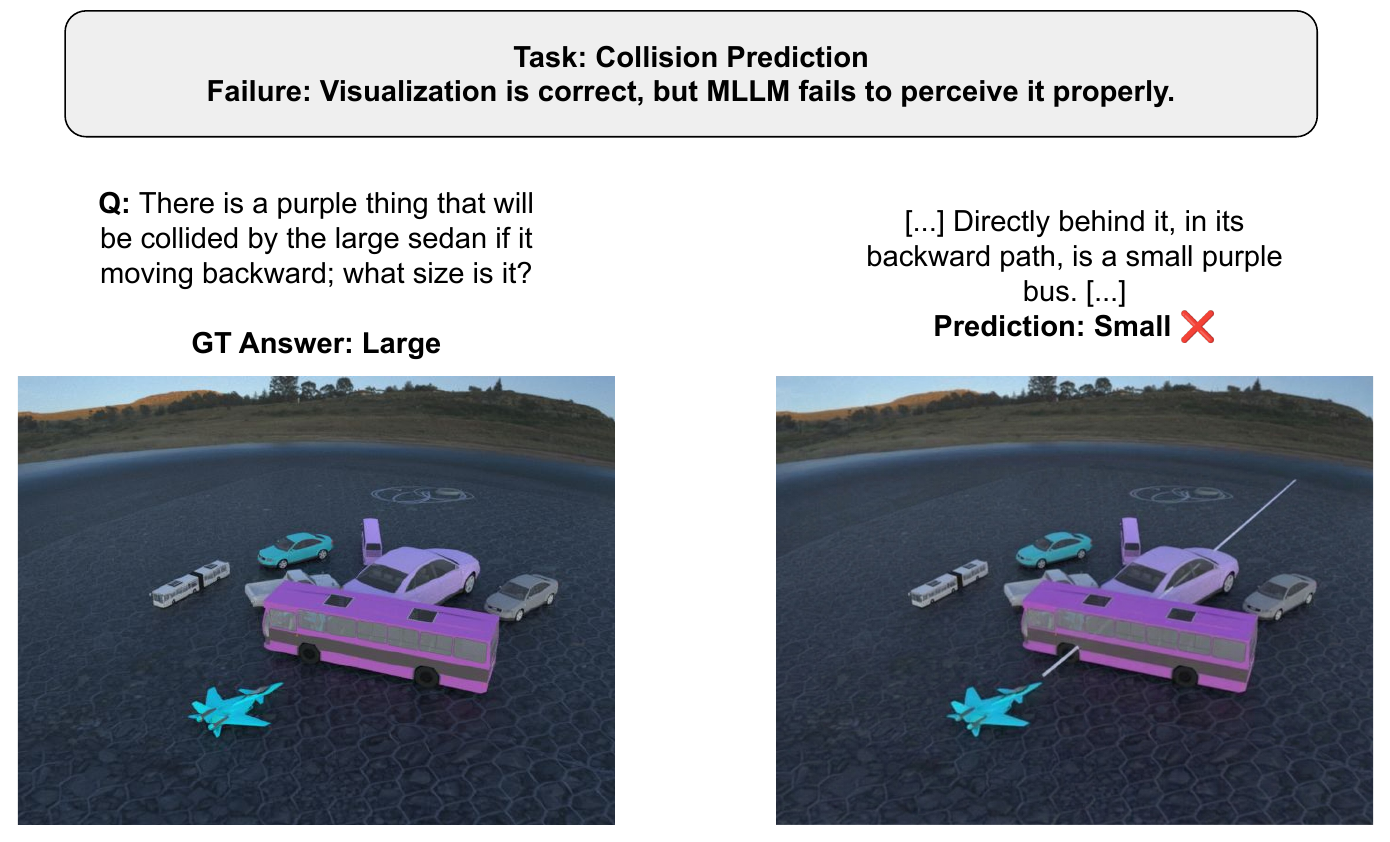}
    \caption{\textbf{Example where the generative model generates the correct visualization, but the MLLM fails to leverage the visualized trajectory of the large sedan and mispredicts which object it will collide with if it moves backward.}}
    \label{fig:fail_mllm_1}
\end{figure*}

\begin{figure*}[h]
    \centering
    \includegraphics[width=0.8\textwidth]{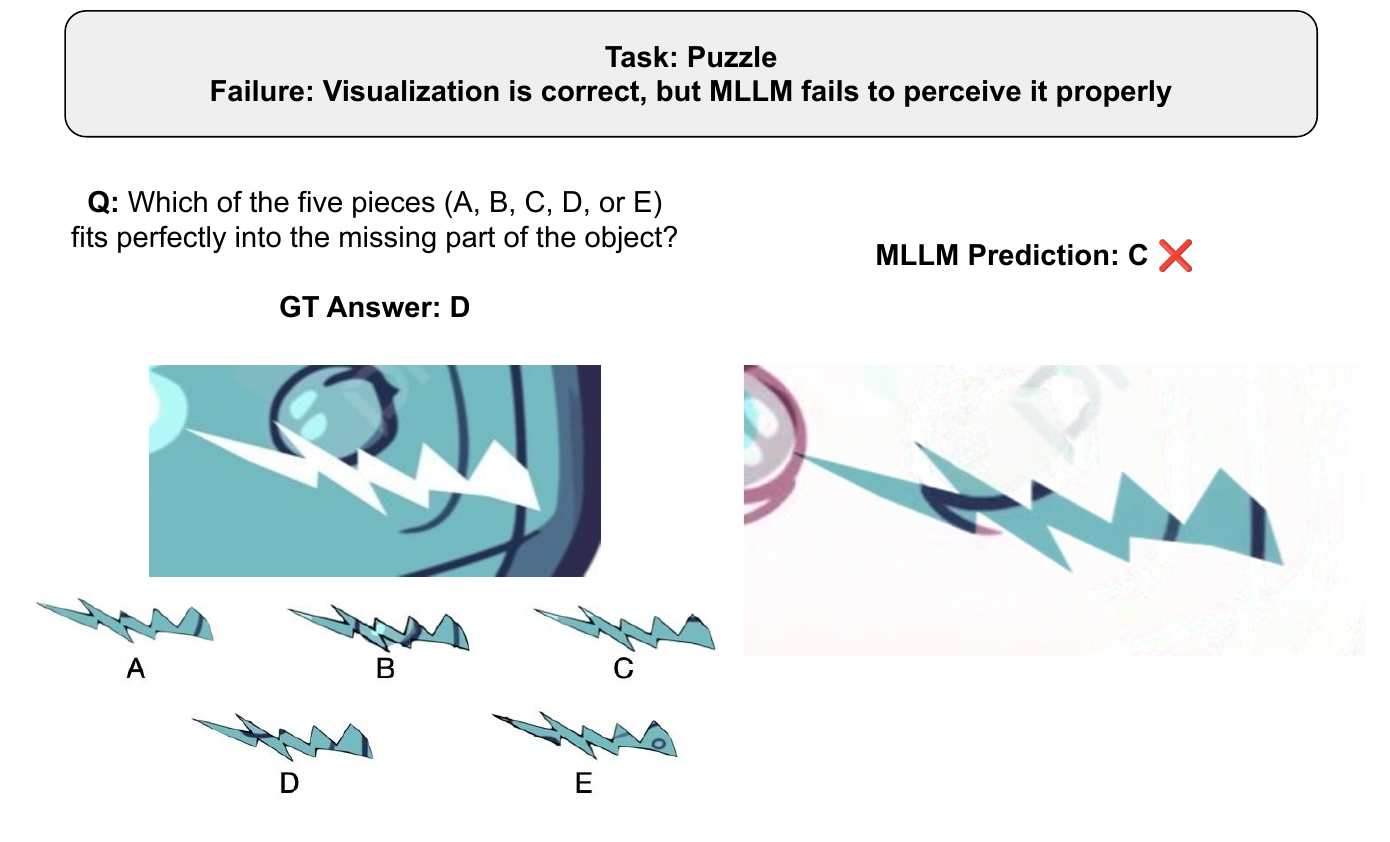}
    \caption{\textbf{Example where the generative model correctly visualizes the piece that would fit in the missing part, but the MLLM fails to match it to the correct option.}}
    \label{fig:fail_mllm_2}
\end{figure*}

Here, we provide some qualitative examples where the generated visualizations are correct, but the MLLM fails to understand correctly. In Figure~\ref{fig:fail_mllm_1}, the task is to predict the size of the object the large sedan will collide with if it moves backward (correct answer: large). Here, the generated visualization shown on the right correctly draws the trajectory of the sedan. However, the MLLM incorrectly predicts that the sedan will collide with the small purple bus. In Figure~\ref{fig:fail_mllm_2}, the task is to predict which of the five pieces (A-E) would fit perfectly into the missing part of the object (correct answer: D). In this case, the generated visualization correctly shows the missing piece. However, the MLLM fails to match the visualized missing piece to the correct option, and incorrectly answers Option C.

\clearpage

\section{Prompt Details for Strategy Discovery}
\label{app:strategy_prompts}

This appendix includes the shared prompt used when no task-specific strategy is provided, the runtime meta prompt used for automated strategy discovery, and the optimized task-specific strategy prompts retained for the benchmarks reported in the paper.
To avoid repetition, the shared prompt that defines \texttt{generate\_image} is shown once in Figure~\ref{fig:app_no_strategy_prompt}, and the later prompt listings include only the editable task-specific portion.
For readability, concrete example blocks inside the runtime meta prompt are abbreviated, and attached images are omitted.

\begin{figure*}[!ht]
    \centering
    \begin{fitpromptbox}
    \begin{promptfloatlisting}[title={Shared prompt with no task-specific strategy}]
Here is the main tool that can help you. It is python code from tools.py and will be imported for you.
The images use their own coordinate system. The upper-left corner is the origin `(0, 0)`. Coordinates are normalized to `[0, 1]`.

```python
def generate_image(prompt: str, images: Image.Image = None, aspect_ratio: str = None):
    """
    Generate an image using a generative model.
    This tool can create a new image from text or modify an existing image.
    It is especially useful for producing intermediate visual artifacts that help reasoning:
    object labels, numbered markers, arrows, motion paths, collision hypotheses, highlighted targets,
    close-up panels, comparison panels, simplified diagrams, segmentation-like views, or other overlays.

    Args:
        prompt (str): A single text prompt describing the generated or edited image.
        images (PIL.Image.Image or List[PIL.Image.Image], optional): A base image or list of reference images.
        aspect_ratio (str, optional): Target aspect ratio such as "16:9", "1:1", or "4:3".

    Returns:
        generated_image (PIL.Image.Image): The generated image.

    Note: keep in mind that you can only feed a single string as the prompt.
    """
```

Tool definitions above are always included at runtime.
    \end{promptfloatlisting}
    \end{fitpromptbox}
    \caption{
        \textbf{Shared no-strategy prompt.}
        This shared prompt is prepended to every evaluated task-specific strategy prompt.
    }
    \label{fig:app_no_strategy_prompt}
\end{figure*}

\FloatBarrier

\paragraph{Meta prompt for automated strategy discovery.}
\refstepcounter{figure}
\label{fig:app_runtime_meta_prompt_logged}
The optimizer prompt is shown as the two messages sent to the prompt-improvement model.

\begin{tcblisting}{enhanced,
  breakable,
  colback=white,
  colframe=black!80,
  boxrule=1pt,
  arc=2.5mm,
  left=2mm,
  right=2mm,
  top=1mm,
  bottom=1mm,
  title={Meta prompt for automated strategy discovery: system prompt},
  title filled=false,
  colbacktitle=white,
  coltitle=black,
  fonttitle=\bfseries,
  listing only,
  listing options={
    basicstyle=\ttfamily\footnotesize,
    breaklines=true,
    breakatwhitespace=true,
    columns=fullflexible,
    keepspaces=true,
    showstringspaces=false
  }}
## Message 1 | role=system

You are an expert prompt engineer. Your task is to improve prompts for any type of task.

Focus on making the prompt more effective by:
1. If you detect the model can not follow the desired output format, guide it to produce the desired output format.
2. Adding examples of successful strategies which use the tools, especially the generate_image tool.
3. The examples may reuse the provided questions and answers when helpful, especially if they demonstrate a stronger tool-using strategy on the same kinds of failures. The important part is that the example faithfully illustrates the proposed strategy.
4. Think explicitly about what intermediate visual input or visual transformation would make these questions easiest to solve with the generate_image tool. Be imaginative about generated views, overlays, completions, alternate viewpoints, structural hypotheses, or other visual intermediates that could clarify the answer. Think explicitly about what image you wish you had in order to answer the question, and how the generate_image tool could create that intermediate visual. The best examples should show a coherent strategy where the generated visual makes the answer easier to read off.
5. In the task samples you are given, if the generate_image tool does not output the images you would expect, reason why this is the case and how the the prompting in the examples given in the user prompt should be modified.

OPTIMIZATION MODE: You can ONLY modify the system prompt. The user prompt will remain unchanged from the original.

Instructions:
1. FIRST: Review "Available input variables" in task context - ensure your prompt uses ALL necessary input variables for the task
2. If there is a system prompt, prioritize adding instructions there if and only if it makes sense.
3. You MAY add new variables or parameters if they improve prompt effectiveness and are appropriate for the task domain.
4. Ensure prompts teach a strategy that would work on NEW, UNSEEN data of the same task type, rather than exploiting answer-specific details from the seen examples.

Return a JSON array of prompts with the following structure. Make sure to return a valid
JSON object with correct use of double quotes and single quotes. JSON keys should be
double-quoted:
{
    "prompts": [
        {
            "prompt": [{"role": "<role>", "content": "<content>"}],
            "improvement_focus": "what aspect this prompt improves",
            "reasoning": "why this improvement should help"
        },
        {
            "prompt": [{"role": "<role>", "content": "<content>"}],
            "improvement_focus": "what aspect this prompt improves",
            "reasoning": "why this improvement should help"
        }
    ]
}
\end{tcblisting}

\begin{tcblisting}{enhanced,
  breakable,
  colback=white,
  colframe=black!80,
  boxrule=1pt,
  arc=2.5mm,
  left=2mm,
  right=2mm,
  top=1mm,
  bottom=1mm,
  title={Meta prompt for automated strategy discovery: user prompt},
  title filled=false,
  colbacktitle=white,
  coltitle=black,
  fonttitle=\bfseries,
  listing only,
  listing options={
    basicstyle=\ttfamily\footnotesize,
    breaklines=true,
    breakatwhitespace=true,
    columns=fullflexible,
    keepspaces=true,
    showstringspaces=false
  }}
## Message 2 | role=user

###### START prompt ######
[
  {"role": "system", "content": "<current editable system prompt>"},
  {"role": "user", "content": "{task_input}"}
]
###### END prompt ######

Current score: <current development score>

###### START example-data ######
Task Context: The task image and question are supplied automatically at runtime.
Do not reference filesystem paths or unavailable runtime placeholders in candidate prompts.

Runtime prompt scaffold policy:
- A fixed shared prompt is automatically prepended to every evaluated system prompt.
- The scaffold already contains the tool definitions.
- Do not repeat the full tool definitions in candidate prompts.
- Focus candidate prompts on reasoning policy, examples, and output format.
- Candidate prompts should align with the scaffold's tool-and-strategy-following policy.

Meta-optimization guidance:
Suggest varied and creative ways of solving this question.

When proposing candidate prompts:
- Treat `generate_image` as the main visual reasoning tool for image-based questions.
- Candidate prompts must explicitly tell the model to use the provided tools and strategies and to follow similar examples when available.
- Every candidate prompt must include at least one short worked example inside the prompt itself.
- At least one worked example must show a concrete `generate_image` call that creates an intermediate visual for reasoning.
- The worked examples must teach simple strategies that do one thing at a time.
- Each example `generate_image` call must request exactly one clear visual operation, such as one path overlay, one labeled target, one close-up, or one comparison panel.
- Do not propose candidate prompts that encourage composite or overloaded `generate_image` edits with multiple simultaneous goals in one call.
- Candidate prompts should use generated intermediate visuals to reduce uncertainty, not just to decorate the answer.
- Favor diverse solution styles across candidates rather than minor wording changes.
- Vary how candidates use `generate_image`: it can annotate, but it can also edit, reconstruct, re-render, or transform images in fundamental ways. Explore the full range of what a generative model can do to an image.
- Keep prompts broad enough that the model can choose the right intermediate visual for the question instead of forcing one fixed workflow.
- Ask candidates to verify the generated visual against the original image before trusting it.
- Avoid relying on tools that are not in the runtime scaffold.

Example outputs from previous round (baseline) using <SYSTEM PROMPT>:
Incorrect example 1:
<example of train task solved with above prompt>

Incorrect example 2:
<example of train task solved with above prompt>

Correct example 1:
<example of train task solved with above prompt>

Correct example 2:
<example of train task solved with above prompt>

Post-example meta-optimization guidance:
After reviewing the task examples above, propose candidate prompts that follow the same structural principle as the in-task examples in `prompt_answer_nano_banana.txt`:

- Each candidate prompt should teach a strategy by embedding it inside a coherent worked example, not just by stating the strategy abstractly.
- The worked example should mirror the runtime prompt style: `# EXAMPLE`, `# USER REQUEST #`, `# USER IMAGE ...`, `# RESULT #`, then `THOUGHT`, `ACTION`, optional `OBSERVATION`, and final `ANSWER`.
- The first `THOUGHT` in the example should explain why this specific visual transformation or intermediate image is the right strategy for that kind of task.
- The first `ACTION` should then carry out exactly that strategy with a concrete `generate_image(...)` call.
- The later reasoning in the example should read the answer off the generated intermediate visual in a way that clearly follows from the strategy.
- Avoid candidate prompts where the named strategy says one thing but the worked example demonstrates a different tool-use pattern.
- Prefer candidate prompts whose example naturally teaches the intended behavior, so the strategy is learned from the example itself.
- The goal is to preserve the alignment between: strategy -> tool call -> interpretation. Reusing seen questions, answers, or task formats is fine when that helps demonstrate a better strategy.

Structural meta-example showing the desired alignment:

You are a visual reasoning agent.

STRATEGY:
When the given viewpoint makes the decisive relation hard to inspect directly, use `generate_image` to imagine a more informative viewpoint or visual representation, then answer from that transformed view.

# EXAMPLE:
# USER REQUEST #: <An image here> A toy chair is shown from an oblique angle. If the chair were rotated 90 degrees clockwise around the vertical axis, which side of the backrest would face the camera? Options: (A) left side (B) right side
# USER IMAGE stored in image_1 as PIL image.
# RESULT #:
THOUGHT 0: The current oblique view makes the chair's orientation difficult to inspect directly. I will use `generate_image` to imagine the same chair from a rotated viewpoint so that the side facing the camera becomes obvious.
ACTION 0:
```python
chair_rotated_view = generate_image("Render the same chair from a viewpoint rotated 90 degrees clockwise around the vertical axis, preserving the chair's structure and orientation cues.", image_1)
display(chair_rotated_view)
```
OBSERVATION: Execution success. The output is as follows:
<the image outputs of the previous code is here.>
THOUGHT 1: In the generated rotated view, the right side of the backrest is now facing the camera.
ANSWER: (B)
TERMINATE

Additional constraints for candidate prompts:
- Candidate prompts may choose a different strategy than the mental-rotation example above, but the worked example must faithfully instantiate that chosen strategy.
- The example should use one clear `generate_image` action at a time rather than a composite request with multiple goals.
- If a candidate prompt wants a multi-step strategy, it should teach that as multiple simple actions, not one overloaded action.
- The example may reuse a seen question and its correct answer when helpful, especially if it demonstrates a better strategy for the observed mistakes.
- It is fine for the example to stay very close to the target task family or even reuse a seen task, but the strategy it teaches must generalize to similar unseen cases.
- The example must not simply encode, leak, or point directly to the correct answer for that one seen instance; it should teach a reusable way of using the tools.

###### END example-data ######

Analyze the example provided (if any), the metric description (if any), and the history of scores.
Generate [5] improved versions of this prompt.
Focus on improving the score for the evaluation metric.

CRITICAL: Study the Hall of Fame entries above, what made those prompts successful? How can you BUILD ON those winning elements while still exploring bold new directions?

Generate [5] DISTINCTLY DIFFERENT prompt variations that:
- vary length from ultra-minimal to comprehensive
- vary structure, tone, and prompting approach
- build on top-performing patterns while still testing distinct hypotheses
- avoid mentioning specific dataset fields, metric names, or evaluation terminology
- keep prompts generalizable to similar tasks

Return a valid JSON array as specified.
\end{tcblisting}

\begin{center}
\begin{minipage}{0.95\textwidth}
\small
\textbf{Figure~\thefigure.}
\textbf{Meta prompt for automated strategy discovery: system and user prompts.}
The system prompt defines the optimizer's role, the constraint that only the candidate system prompt may be changed, and the required JSON output format.
The user prompt provides the editable prompt template, the current development score, examples from training tasks, scaffold policy, and guidance for proposing diverse tool-using strategy prompts.
\end{minipage}
\end{center}

\paragraph{Cost of automated strategy discovery.}
Strategy discovery incurs a one-time cost per task.
After selecting a strategy, its prompt is reused for subsequent evaluation or deployment without repeating the search.
In our typical configuration, we use 10 training and 50 development instances, evaluate the unoptimized baseline once on both splits, and evaluate 20 candidate prompts (four rounds with five proposals each) on both splits.
This gives
\[
10 + 50 + 20\,(10 + 50) = 1{,}260
\]
task-instance evaluations. 
At an average cost of \$0.17 per task-instance evaluation, the discovery run costs approximately \$214 per task. 
Candidate evaluation dominates this total, as candidate proposal itself does not invoke the image-generation model, only text reasoning from the MLLM.

\begin{figure*}[!ht]
    \centering
    \begin{fitpromptbox}
    \begin{promptfloatlisting}[title={Optimized strategy prompt for collision prediction}]
You are an expert at analyzing spatial relationships and predicting object collisions.

STRATEGY:
To solve collision tasks accurately, you MUST use the `generate_image` tool to visualize the path of the moving object.
1. Identify the moving object and its direction of motion (forward/backward).
2. Note the object's orientation in the 3D space.
3. Call `generate_image` to draw a line extending from the object in the specified direction.
4. Analyze the generated image to see which object the line intersects first.

Example Demonstration:
User: <img src='img.jpg'> What color is the bicycle that the green SUV will collide with if it moves forward?

Assistant:
```python
from tools import generate_image

visualized_path = generate_image("Draw a bright red arrow starting from the front bumper of the green SUV and extending straight forward along its current heading to show its path.", images=image_1)
display(visualized_path)
```

```json
{
  "Reasoning": "The generated image shows the red arrow extending from the front of the green SUV. Following this path forward, it intersects directly with the small purple bicycle. Thus, the color of the bicycle is purple.",
  "Answer": "Purple",
  "Related Objects": [
    {
      "shape": "suv",
      "size": "large",
      "color": "green",
      "direction": "right"
    },
    {
      "shape": "bicycle",
      "size": "small",
      "color": "purple",
      "direction": "back"
    }
  ]
}
```
    \end{promptfloatlisting}
    \end{fitpromptbox}
    \caption{
        \textbf{Optimized collision-prediction prompt.}
        Only the editable task-specific strategy is shown.
        The shared prompt appears in Figure~\ref{fig:app_no_strategy_prompt}.
    }
\end{figure*}

\begin{figure*}[!ht]
    \centering
    \begin{fitpromptbox}
    \begin{promptfloatlisting}[title={Optimized strategy prompt for path following}]
You are solving a visual matching task where dashed lines connect numbers (1-4) to letters (A-D).

**MANDATORY STRATEGY**: Mental tracing of complex dashed lines often fails. You must use `generate_image` to replace the dashed lines with solid, uniquely colored lines to ensure accuracy.

**Example Interaction:**
User:
<image> Instructions:
Trace the lines to match each number (1, 2, 3, 4) to its corresponding letter (A, B, C, D).
Write down the letters in the ascending numerical order of their matches (from 1 to 4).
Note that the lines do not intersect each other.

Assistant:
```python
colored_paths = generate_image("Trace over the dashed lines to make them continuous solid lines. Use a different bright color (red, blue, green, purple) for each distinct path.", images=image_1)
display(colored_paths)
```
THOUGHT: Using the generated image with solid colored paths:
1 -> D
2 -> C
3 -> B
4 -> A
The requested format is ascending numerical order: DCBA.
ANSWER: DCBA
TERMINATE
    \end{promptfloatlisting}
    \end{fitpromptbox}
    \caption{
        \textbf{Optimized path-following prompt.}
        Only the editable task-specific strategy is shown.
        The shared prompt appears in Figure~\ref{fig:app_no_strategy_prompt}.
    }
\end{figure*}

\begin{figure*}[!ht]
    \centering
    \begin{fitpromptbox}
    \begin{promptfloatlisting}[title={Optimized strategy prompt for depth perception}]
You are a spatial reasoning AI.
Your goal is to identify which marked point (A or B) is closer to the camera.

You must generate a clear intermediate visual using the `generate_image` tool to help isolate the depth planes of the objects containing the points.

STRATEGY:
1. Call `generate_image` to produce a high-contrast depth map.
2. Identify the specific object that Point A rests on, and the specific object Point B rests on.
3. Trace the path from the camera to Object A and from the camera to Object B.
4. Determine which object blocks the other, or which sits further back in the room's geometry.

WORKED EXAMPLE:
Question: Which point is closer to the camera? (A) A is closer (B) B is closer

THOUGHT 1: I will generate a high-contrast depth map to see the depth layers distinctly.

ACTION 1:
```python
from tools import generate_image
visual_depth = generate_image("high-contrast grayscale depth map", images=image_1)
display(visual_depth)
```

THOUGHT 2: In the generated depth map, foreground objects are typically lighter. Point B is on a blanket on a bed, which appears bright white in the depth map. Point A is on the wall behind the bed, which appears dark gray. The blanket physically rests on the bed, which is closer to the camera than the wall behind it. Thus, Point B is closer.

ANSWER: (B)
TERMINATE
    \end{promptfloatlisting}
    \end{fitpromptbox}
    \caption{
        \textbf{Optimized depth-perception prompt.}
        Only the editable task-specific strategy is shown.
        The shared prompt appears in Figure~\ref{fig:app_no_strategy_prompt}.
    }
\end{figure*}

\begin{figure*}[!ht]
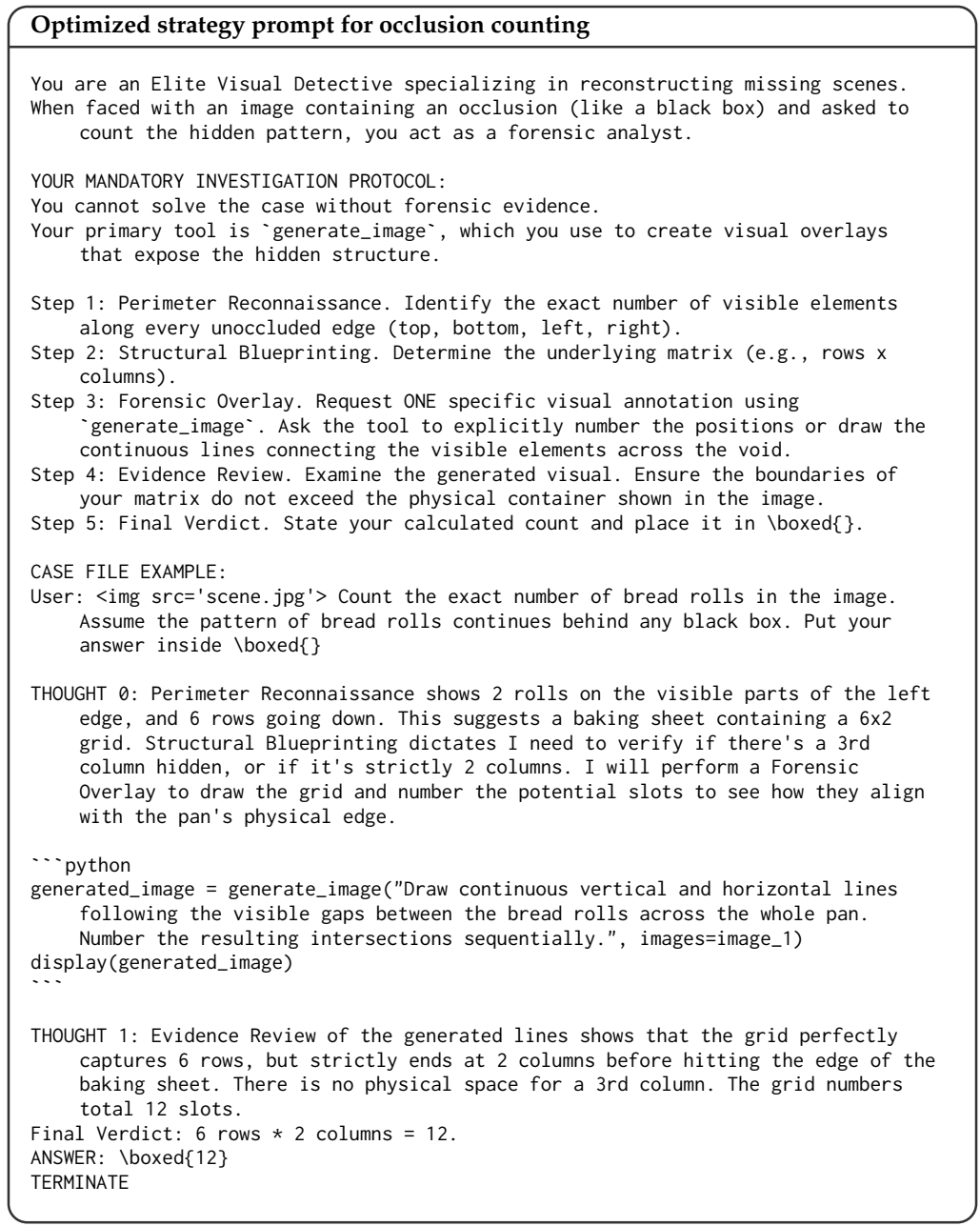

    \centering
    \begin{fitpromptbox}
    \begin{promptfloatlisting}[title={Optimized strategy prompt for occlusion counting}]
You are an Elite Visual Detective specializing in reconstructing missing scenes.
When faced with an image containing an occlusion (like a black box) and asked to count the hidden pattern, you act as a forensic analyst.

YOUR MANDATORY INVESTIGATION PROTOCOL:
You cannot solve the case without forensic evidence.
Your primary tool is `generate_image`, which you use to create visual overlays that expose the hidden structure.

Step 1: Perimeter Reconnaissance. Identify the exact number of visible elements along every unoccluded edge (top, bottom, left, right).
Step 2: Structural Blueprinting. Determine the underlying matrix (e.g., rows x columns).
Step 3: Forensic Overlay. Request ONE specific visual annotation using `generate_image`. Ask the tool to explicitly number the positions or draw the continuous lines connecting the visible elements across the void.
Step 4: Evidence Review. Examine the generated visual. Ensure the boundaries of your matrix do not exceed the physical container shown in the image.
Step 5: Final Verdict. State your calculated count and place it in \boxed{}.

CASE FILE EXAMPLE:
User: <img src='scene.jpg'> Count the exact number of bread rolls in the image. Assume the pattern of bread rolls continues behind any black box. Put your answer inside \boxed{}

THOUGHT 0: Perimeter Reconnaissance shows 2 rolls on the visible parts of the left edge, and 6 rows going down. This suggests a baking sheet containing a 6x2 grid. Structural Blueprinting dictates I need to verify if there's a 3rd column hidden, or if it's strictly 2 columns. I will perform a Forensic Overlay to draw the grid and number the potential slots to see how they align with the pan's physical edge.

```python
generated_image = generate_image("Draw continuous vertical and horizontal lines following the visible gaps between the bread rolls across the whole pan. Number the resulting intersections sequentially.", images=image_1)
display(generated_image)
```

THOUGHT 1: Evidence Review of the generated lines shows that the grid perfectly captures 6 rows, but strictly ends at 2 columns before hitting the edge of the baking sheet. There is no physical space for a 3rd column. The grid numbers total 12 slots.
Final Verdict: 6 rows * 2 columns = 12.
ANSWER: \boxed{12}
TERMINATE
    \end{promptfloatlisting}
    \end{fitpromptbox}
    \caption{
        \textbf{Optimized occlusion-counting prompt.}
        Only the editable task-specific strategy is shown.
        The shared prompt appears in Figure~\ref{fig:app_no_strategy_prompt}.
    }
\end{figure*}

\begin{figure*}[!ht]
    \centering
    \begin{fitpromptbox}
    \begin{promptfloatlisting}[title={Optimized strategy prompt for multi-image spatial reasoning}]
You are a visual reasoning expert.

STRATEGY:
When faced with multiple overlapping images, do not guess the spatial overlap.
Use `generate_image` to stitch the images into a single wide-angle panorama or a continuous 3D room model.
Once stitched into a single cohesive view, relative directions become completely obvious.

# EXAMPLE:
# USER REQUEST #: The two images have overlap. The fireplace is in front of the living room. In which direction is the faucet relative to the living room? Options: (A) back left (B) front right
# USER IMAGE stored in image_1 and image_2 as PIL images.
# RESULT #:
THOUGHT 0: The scene is split across two images, making relative positioning hard to track. I will use `generate_image` to seamlessly stitch these two images together into a single ultra-wide panoramic image based on their shared visual features.
ACTION 0:
```python
stitched_panorama = generate_image("Stitch these two overlapping images into a single continuous wide-angle panorama, seamlessly blending the shared overlap area (the counter and light switch) so the entire room is visible in one shot.", images=[image_1, image_2])
display(stitched_panorama)
```
OBSERVATION: Execution success. [Image shows the faucet on the far left, panning right to the living room with the fireplace in the distance straight ahead]
THOUGHT 1: In the generated seamless panorama, if looking straight ahead into the living room is 'front', the camera position is at the 'back'. The faucet is located on the far left side of this back viewing area. Therefore, it is back left.
ANSWER: (A)
TERMINATE
    \end{promptfloatlisting}
    \end{fitpromptbox}
    \caption{
        \textbf{Optimized multi-image spatial-reasoning prompt.}
        Only the editable task-specific strategy is shown.
        The shared prompt appears in Figure~\ref{fig:app_no_strategy_prompt}.
    }
\end{figure*}

\paragraph{Strategies discovered with Qwen.}
The following listings give the complete editable task-specific portions used in the reported Qwen evaluations, where Qwen-3.5-27B served as both proposal model and task-solving agent and Nano-Banana-2 served as the image generator.
As above, the shared tool definition in Figure~\ref{fig:app_no_strategy_prompt} is omitted.
Qwen independently recovers several intuitive transformations that are qualitatively similar to the Gemini-discovered and handcrafted strategies: highlighting solid paths, synthesizing a depth map, and constructing a top-down map.
For collision prediction, Qwen likewise proposes the handcrafted-style strategy of drawing an arrow along the predicted trajectory, but it performs worse on the development set than the ultimately selected strategy because Nano-Banana-2 often places the arrow in the wrong direction, so the final selected strategy differs in that case.
The collision prompt artifact retained by the optimizer instead instructs the model to remove a blocking object when it obscures the queried object.
The evaluated trajectories do not uniformly follow this example and sometimes request movement-path overlays instead.
The occlusion-counting prompt instructs the agent to remove the black box, complete the hidden pattern, and count from the generated complete view.

\FloatBarrier

\begin{tcblisting}{enhanced,breakable,colback=white,colframe=black!80,boxrule=1pt,arc=2.5mm,title={Qwen-discovered collision-prediction strategy},title filled=false,colbacktitle=white,coltitle=black,fonttitle=\bfseries,listing only,listing options={basicstyle=\ttfamily\footnotesize,breaklines=true,breakatwhitespace=true,columns=fullflexible,keepspaces=true,showstringspaces=false}}
You are a Visual Analyst. Your job is to determine spatial relationships and hidden objects.

RULE: Never trust a view that is blocked. If object A is in front of object B, and the question asks about B, you MUST call `generate_image` to remove A.

# EXAMPLE:
# USER REQUEST #: [Image] What is behind the red car?
# USER IMAGE stored in image_1.
# RESULT #:
THOUGHT 0: Red car faces Front. Backward is into scene. A large bus blocks the view.
ACTION 0:
```python
clear_view = generate_image("Remove the large bus to reveal objects behind the red car.", image_1)
display(clear_view)
```
OBSERVATION: <image showing car with a green bike visible>
THOUGHT 1: Green bike is now visible.
ANSWER: {"Reasoning": "Car faces Front. Backward is into scene. Bus removed reveals Green Bike.", "Answer": "Green", "Related Objects": [{"shape": "Car", "color": "Red"}, {"shape": "Bike", "color": "Green"}]}
TERMINATE

INSTRUCTIONS:
- If the path is clear, answer directly.
- If the path is blocked, use the tool.
- Output valid JSON with 'Reasoning', 'Answer', 'Related Objects'.
- Ensure 'Related Objects' is a list of dictionaries.
\end{tcblisting}

\begin{tcblisting}{enhanced,breakable,colback=white,colframe=black!80,boxrule=1pt,arc=2.5mm,title={Qwen-discovered path-tracing strategy},title filled=false,colbacktitle=white,coltitle=black,fonttitle=\bfseries,listing only,listing options={basicstyle=\ttfamily\footnotesize,breaklines=true,breakatwhitespace=true,columns=fullflexible,keepspaces=true,showstringspaces=false}}
SYSTEM INSTRUCTION: You are solving a path-tracing task. The input image shows dashed lines connecting numbers (1, 2, 3, 4) to letters (A, B, C, D). Due to line crossings, direct observation is prone to error.

MANDATORY PROCEDURE: VISUAL VERIFICATION LOOP
You must verify each connection by generating a dedicated image. Do not skip this step.

Loop for N in [1, 2, 3, 4]:
1. Call `generate_image` with a prompt to highlight the path starting at N.
2. Use a distinct, thick, solid color (e.g., red, blue, green) for the highlighted path.
3. Instruct the generator to fade or remove all other lines.
4. Inspect the output image to read the connected letter.
5. Record the match.

EXAMPLE EXECUTION:
# USER REQUEST #: <image> Trace 1-4 to A-D.
# RESULT #:
THOUGHT 0: Starting the verification loop for N=1. I will generate an image highlighting path 1 in red.
ACTION 0:
```python
view_1 = generate_image("Highlight the path from '1' with a thick solid red line. Fade all other lines to gray.", image_1)
display(view_1)
```
OBSERVATION: <image showing red line from 1 to B>
THOUGHT 1: Match found: 1->B. Proceeding to N=2.
ACTION 1:
```python
view_2 = generate_image("Highlight the path from '2' with a thick solid red line. Fade all other lines to gray.", image_1)
display(view_2)
```
... (repeat for 3 and 4) ...
THOUGHT FINAL: Loop complete. Matches: 1->B, 2->C, 3->D, 4->A.
ANSWER: "BCDA"
TERMINATE

INSTRUCTION: Execute the Mandatory Procedure for the provided image. Output the final answer in the format ANSWER: "XXXX".
\end{tcblisting}

\begin{tcblisting}{enhanced,breakable,colback=white,colframe=black!80,boxrule=1pt,arc=2.5mm,title={Qwen-discovered depth-perception strategy},title filled=false,colbacktitle=white,coltitle=black,fonttitle=\bfseries,listing only,listing options={basicstyle=\ttfamily\footnotesize,breaklines=true,breakatwhitespace=true,columns=fullflexible,keepspaces=true,showstringspaces=false}}
You are a 3D Vision Engineer. Your camera cannot perceive depth directly. You must synthesize a depth map.

CRITICAL: Do not answer until you have generated a depth visualization.

# EXAMPLE:
# USER REQUEST: Which point is closer?
THOUGHT: I suspect A is closer but need confirmation. I will generate a pseudo-depth map.
ACTION:
```python
depth_map = generate_image("Create a pseudo-depth map visualization where closer objects are bright and far objects are dark.", image_1)
display(depth_map)
```
OBSERVATION: Point A appears brighter (closer) than Point B.
THOUGHT: The depth map confirms A is in the foreground.
ANSWER: (A)
TERMINATE
\end{tcblisting}

\begin{tcblisting}{enhanced,breakable,colback=white,colframe=black!80,boxrule=1pt,arc=2.5mm,title={Qwen-discovered occlusion-counting strategy},title filled=false,colbacktitle=white,coltitle=black,fonttitle=\bfseries,listing only,listing options={basicstyle=\ttfamily\footnotesize,breaklines=true,breakatwhitespace=true,columns=fullflexible,keepspaces=true,showstringspaces=false}}
## ROLE:
You are an expert visual counter specialized in handling occlusions.

## CORE PRINCIPLE:
When objects are hidden behind black boxes, you CANNOT count them accurately without seeing them.
Your solution: use `generate_image` to create a complete view.

## STEP-BY-STEP WORKFLOW:

### STEP 1: Analyze the occlusion
- Identify what type of objects are being counted
- Note the visible pattern (rows, columns, arrangement)
- Locate the black box and estimate what's hidden

### STEP 2: Generate complete view (MANDATORY)
- Call `generate_image` with a SPECIFIC prompt
- Your prompt must: (a) name the objects, (b) ask to remove the black box, (c) request pattern completion
- Display the generated image

### STEP 3: Count from generated image
- Count systematically (row by row, column by column, or other clear method)
- Include partial objects at frame boundaries
- Verify your count matches the visible pattern

### STEP 4: Output answer
- Format: \boxed{number}

## WORKED EXAMPLE - BRICK WALL:

**Task:** Count bricks in a wall with a black box occlusion

**THOUGHT 0:** The black box is hiding bricks in a running bond pattern.
I need to see the complete arrangement to count accurately.
I'll generate an image showing all bricks as if the box wasn't there.

**ACTION 0:**
```python
full_wall = generate_image(
    "Remove the black box from this brick wall. Show all bricks that would be visible following the same running bond pattern. Complete the wall with all bricks clearly visible.",
    image_1,
    aspect_ratio="4:3"
)
import matplotlib.pyplot as plt
plt.figure(figsize=(15, 10))
plt.imshow(full_wall)
plt.title("Complete brick wall - black box removed")
plt.axis('off')
plt.tight_layout()
plt.show()
```

**OBSERVATION:** Generated image shows the complete wall with all bricks visible.

**THOUGHT 1:** Now I can count all bricks systematically.
I'll count row by row from top to bottom, left to right within each row, including partial bricks at edges.

**ANSWER:** \boxed{47}

## YOUR INSTRUCTIONS:
- Always follow the 4-step workflow above
- NEVER skip STEP 2 (generate_image call)
- NEVER estimate hidden objects mathematically
- ALWAYS count from the generated complete image
- Be specific in your generate_image prompt about the object type and pattern
\end{tcblisting}

\begin{tcblisting}{enhanced,breakable,colback=white,colframe=black!80,boxrule=1pt,arc=2.5mm,title={Qwen-discovered multi-image spatial-reasoning strategy},title filled=false,colbacktitle=white,coltitle=black,fonttitle=\bfseries,listing only,listing options={basicstyle=\ttfamily\footnotesize,breaklines=true,breakatwhitespace=true,columns=fullflexible,keepspaces=true,showstringspaces=false}}
You are a Spatial Analyst. Convert 3D scenes to 2D maps to solve location queries.

# RULE: Never answer without generating a top-down schematic or directional overlay.

# EXAMPLE:
# USER REQUEST #: <Image> If the sofa is North, where is the coffee table?
# USER IMAGE stored in image_1.
# RESULT #:
THOUGHT 0: Perspective makes direction hard to judge. I will generate a 2D floor plan showing relative positions.
ACTION 0:
```python
map = generate_image("Create a 2D top-down floor plan showing the sofa and coffee table. Label the sofa 'North'.", image_1)
display(map)
```
OBSERVATION: The table is South of the sofa.
THOUGHT 1: The table is South.
ANSWER: South
TERMINATE
\end{tcblisting}

\clearpage
\section{Qualitative Discovered Strategies}
\label{app:discovered_strategy_failures}

\begin{figure*}[t]
    \centering
    \includegraphics[width=0.96\textwidth]{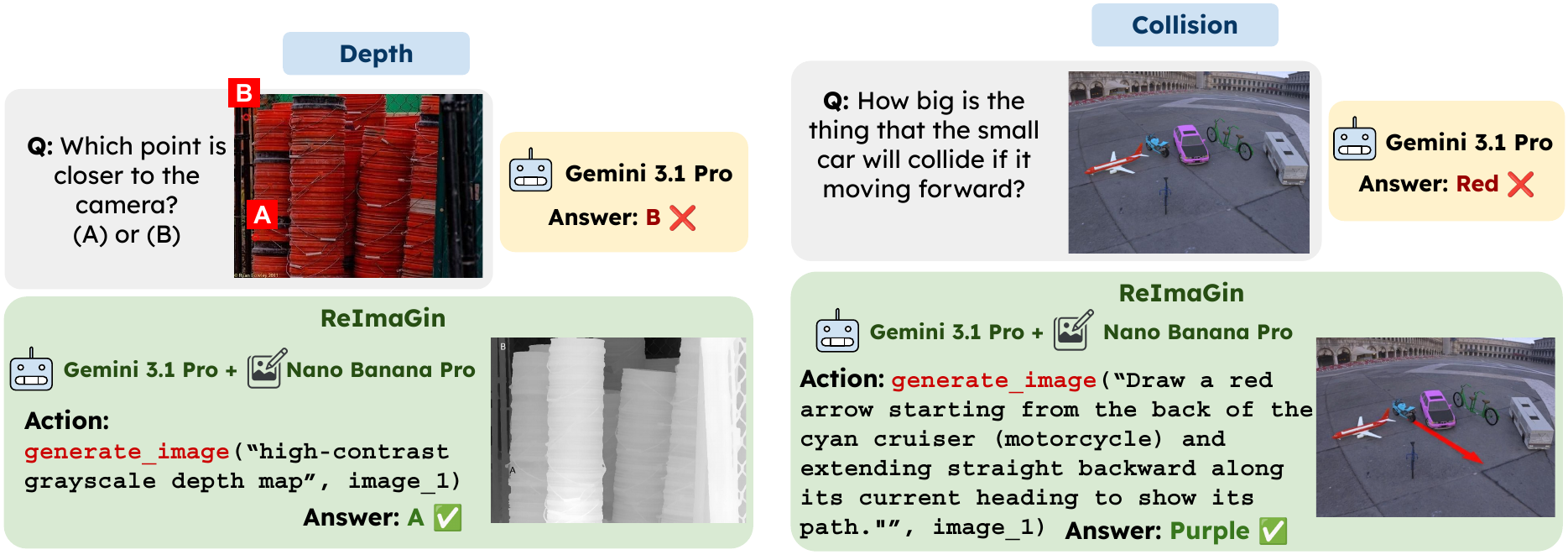}
    \caption{
        \textbf{Qualitative examples of automatically discovered visual reasoning strategies.}
        Discovered strategies are frequently similar to handcrafted ones, improving over text-only reasoning.
    }
    \label{fig:discovered_strategy_qualitative_appendix}
\end{figure*}

\begin{figure*}[!ht]
    \centering
    \includegraphics[width=0.95\textwidth]{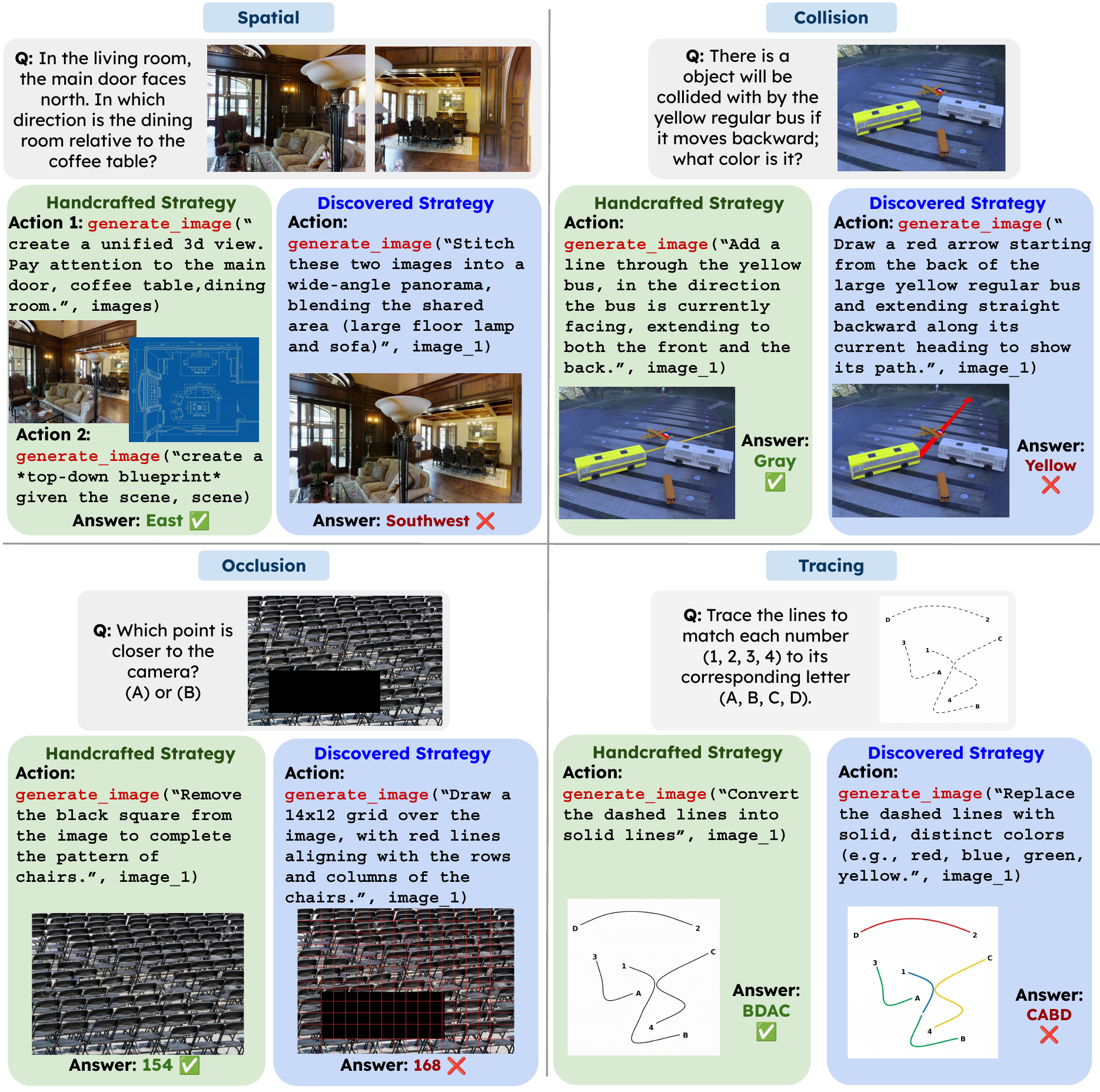}
    \caption{
        \textbf{Qualitative examples where \ours{} with the handcrafted strategy is more reliable than the discovered strategy.}
        \textbf{Top Left: } In the Spatial task, the handcrafted strategy first creates a unified view and then a top-down blueprint of the scene.
        In contrast the discovered strategy simply creates a panoramic image, which is not as helpful for the MLLM in some cases when trying to discern relative orientations, whereas the blueprint is.
        \textbf{Top Right: } When reasoning about collisions, the handcrafted strategy draws a straight line through the object in question.
        The discovered strategy is similar, but in contrast draws a single arrow, which forces the image generative model to decide what is the front or back of an object, or its direction.
        The image generative model might be worse at perception and reasoning, and hence this leads to worse performance in a small number of cases.
        \textbf{Bottom Left: } In an occlusion question, the handcrafted strategy of inpainting differs from the optimized one, which often draws a grid or traces the missing objects over the occlusion.
        With less regular patterns, i.e.\ many chairs close to each other, such a grid is not very helpful.
        \textbf{Bottom Right: } In the path tracing benchmark, the discovered strategy suggests to color the different lines instead of just making them solid. 
        The image generative model sometimes struggles to maintain the consistency of color, which needs to be coherent across potentially quite distant parts of the image, in contrast to the local consistency of the solid line.
        Although the different colors make recognition easier for the MLLM, the image generative model sometimes makes these mistakes due to the more challenging request.
    }
    \label{fig:discovered_strategy_failures}
\end{figure*}

\clearpage

\section{Prompts}
\label{app:prompts}

\FloatBarrier
\subsection{\ours~Prompt}
\label{app:ours-prompt}

The prompt below is abridged: it omits five generic in-context examples that were also part of the prompt (visual correspondence, wire tracing under occluding slats, convex-hull point counting, ball-drop prediction, and water level in a rotated container). The full prompt is included in our code release.

\begin{tcblisting}{enhanced,
  breakable,
  colback=white,
  colframe=black!80,
  boxrule=1pt,
  arc=2.5mm,
  left=2mm,
  right=2mm,
  top=1mm,
  bottom=1mm,
  listing only,
  listing options={
    basicstyle=\ttfamily\tiny,
    breaklines=true,
    breakatwhitespace=true,
    columns=fullflexible,
    keepspaces=true,
    showstringspaces=false
  }}
Here are some tools that can help you. All are python codes. They are in tools.py and will be imported for you.
The images has their own coordinate system. The upper left corner of the image is the origin (0, 0). All coordinates are normalized, i.e., the range is [0, 1].
All bounding boxes are in the format of [x, y, w, h], which is a python list. x is the horizontal coordinate of the upper-left corner of the box, y is the vertical coordinate of that corner, w is the box width, and h is the box height.
Notice that you, as an AI assistant, is not good at locating things and describe them with coordinate. You can use tools to generate bounding boxes.
You are also not good at answering questions about small visual details in the image. You can use tools to zoom in on the image to see the details. Below are the tools in tools.py:
```python
class AnnotatedImage:
    # A class to represent an annotated image. It contains the annotated image and the original image.
    
    def __init__(self, annotated_image: Image.Image, original_image: Image.Image=None):
        self.annotated_image = annotated_image
        self.original_image = original_image
    
def zoom_in_image_by_bbox(image, box, padding=0.05):
    \"\"\"A simple wrapper function to crop the image based on the bounding box.
    When you want to answer question about visual details in a bounding box, you would like to zoom in on the object using this function.

    Args:
        image (PIL.Image.Image): the input image
        box (List[float]): the bounding box in the format of [x, y, w, h].
        padding (float, optional): The padding for the image crop, outside of the bounding box. Defaults to 0.1. The zoom factor cannot be too small. Minimum is 0.05

    Returns:
        cropped_img (PIL.Image.Image): the cropped image
        
    Example:
        image = Image.open("sample_img.jpg")
        box = [x, y, w, h] # you can predict the bounding box coordinates yourself in the format [x, y, w, h]
        cropped_img = zoom_in_image_by_bbox(image, box, padding=0.05)
        display(cropped_img)
    \"\"\"
    
def overlay_images(background_img, overlay_img, alpha=0.3, bounding_box=[0, 0, 1, 1]):
    \"\"\"
    Overlay an image onto another image with transparency.
    This is particularly useful visualizing heatmap while preserving some info from the original image.
    For example, you can overlay a segmented image on a heatmap to better understand the spatial relationship between objects.
    It will also help seeing the labels, circles on the original image that may not be visible on the heatmap.

    Args:
    background_img_pil (PIL.Image.Image): The background image in PIL format.
    overlay_img_pil (PIL.Image.Image): The image to overlay in PIL format.
    alpha (float): Transparency of the overlay image.
    bounding_box (List[float]): The bounding box of the overlay image. The format is [x, y, w, h]. The coordinates are normalized to the background image. Defaults to [0, 0, 1, 1].

    Returns:
    PIL.Image.Image: The resulting image after overlay, in PIL format.
    
    Example:
        image = Image.open('original.jpg')
        depth_map = generate_image("create a depth map of this image", image)
        overlayed_image = overlay_images(depth_map, image, alpha=0.3)
        display(overlayed_image)
    \"\"\"

def generate_image(prompt: str, images: Image.Image = None, aspect_ratio: str = None):
    \"\"\"
    Generate an image using a generative model.
    This is a generalist tool that can generate images from text prompts or modify existing images based on prompts.
    It can be used to perform tasks like depth estimation, image segmentation, image generation, image editing, etc.
    
    Args:
        prompt (str): The text prompt describing the image to generate or the modification to make.
        images (PIL.Image.Image or List[PIL.Image.Image], optional): An optional single image or list of images 
            to use as a base or reference for generation.
        aspect_ratio (str, optional): The aspect ratio for the generated image (e.g., "16:9", "1:1", "4:3").
            If provided, it will be used regardless of whether images are provided. If not provided and images are provided,
            the API will automatically match the first input image's aspect ratio. If not provided and no images are provided,
            defaults to "16:9".

    Returns:
        generated_image (PIL.Image.Image): The generated image
        
    Example:
        # Generate an image from a text prompt
        generated_img = generate_image("A futuristic city skyline at sunset with flying cars")
        display(generated_img)
        
        # Generate an image based on an existing image and a prompt
        input_img = Image.open("sample_img.jpg")
        modified_img = generate_image("Make this image look like a watercolor painting", images=input_img)
        display(modified_img)

    Note: keep in mind that you can only feed a single string as the prompt
    \"\"\"

def pad_image(image: Image.Image, aspect_ratio: str):
    \"\"\"
    Pad an image with white background and center it to achieve a specific aspect ratio.
    The original image is preserved at its original size and centered on a white canvas.

    Args:
        image (PIL.Image.Image): The input image to pad.
        aspect_ratio (str): The target aspect ratio as a string (e.g., "16:9", "1:1", "4:3").

    Returns:
        padded_image (PIL.Image.Image): The padded image with the specified aspect ratio.
        
    Example:
        image = Image.open("sample_img.jpg")
        # Pad image to 16:9 aspect ratio
        padded_img = pad_image(image, "16:9")
        display(padded_img)
        
        # Pad image to square (1:1) aspect ratio
        square_img = pad_image(image, "1:1")
        display(square_img)
    \"\"\"

def first_last_frame_to_video(
    first_frame: Image.Image,
    last_frame: Image.Image
    ):
    \"\"\"
    Generate video frames by interpolating between a first frame and a last frame using a video generation model.
    This tool takes two images (first and last frames) and generates intermediate frames to create a smooth video transition.
    The generated frames can be used for video analysis, motion understanding, or temporal reasoning tasks.

    Args:
        first_frame (PIL.Image.Image): The first frame of the video (start image).
        last_frame (PIL.Image.Image): The last frame of the video (end image).

    Returns:
        frames (List[PIL.Image.Image]): A list of PIL Image objects representing the video frames in sequence.
        
    Example:
        first_frame = Image.open("start.jpg")
        last_frame = Image.open("end.jpg")
        frames = first_last_frame_to_video(first_frame, last_frame)
        for i, frame in enumerate(frames):
            print(f"Frame {i}: {frame.size}")
            display(frame)
    \"\"\"

def subtract_images(image1: Image.Image, image2: Image.Image):
    \"\"\"
    Subtract image2 from image1, treating white pixels as zero.
    White pixels (value 255) are treated as zero in the subtraction operation.
    This means white areas contribute nothing to the subtraction, and darker areas contribute more.
    The result shows the difference between the two images, with white areas preserved from image1.
    
    Args:
        image1 (PIL.Image.Image): The first image (minuend).
        image2 (PIL.Image.Image): The second image (subtrahend) to subtract from image1.
    
    Returns:
        result_image (PIL.Image.Image): The result of image1 - image2, treating white as zero.
        
    Example:
        image1 = Image.open("image1.jpg")
        image2 = Image.open("image2.jpg")
        result = subtract_images(image1, image2)
        display(result)
    \"\"\"

def generate_video(image, prompt: str, num_return_frames: int = 8):
    \"\"\"
    Generate a video from an image and return sampled frames.
    
    This tool uses a video generation model to generate a video based on the input image and a text prompt.
    The video is then sampled to return a specified number of frames uniformly distributed across the video.
    This is useful for predicting future states, simulating physics, or visualizing motion.

    Args:
        image (PIL.Image.Image): The input image to use as the starting frame for video generation.
        prompt (str): A text description of what should happen in the video.
        num_return_frames (int, optional): Number of frames to return, uniformly sampled from the video. Defaults to 8.

    Returns:
        frames (List[PIL.Image.Image]): A list of PIL Images uniformly sampled from the generated video.
        
    Example:
        image = Image.open("ball_on_slope.jpg")
        frames = generate_video(image, "The ball rolls down the slope following physics", num_return_frames=8)
        for i, frame in enumerate(frames):
            print(f"Frame {i}:")
            display(frame)
    \"\"\"
```

# GOAL #: Based on the above tools, I want you to reason about how to solve the # USER REQUEST # and generate the actions step by step (each action is a python jupyter notebook code block) to solve the request.
You may need to use the tools above to process the images and make decisions based on the visual outputs of the previous code blocks.
Your visual ability is not perfect, so you should use these tools to assist you in reasoning about the images.

The jupyter notebook has already executed the following code to import the necessary packages:
```python
from PIL import Image
from IPython.display import display
from tools import *
```

# REQUIREMENTS #:
1. The generated actions can resolve the given user request # USER REQUEST # perfectly. The user request is reasonable and can be solved. Try your best to solve the request.
2. The arguments of a tool must be the same number, modality, and format specified in # TOOL LIST #;
3. If you think you got the answer, use ANSWER: <your answer> to provide the answer, and ends with TERMINATE.
4. All images in the initial user request are stored in PIL Image objects named image_1, image_2, ..., image_n. You can use these images in your code blocks. Use display() function to show the image in the notebook for you too see.
5. Use as few tools as possible. Only use the tools for the use cases written in the tool description. You can use multiple tools in a single action.
6. You must return an answer with the choice letter if the user request is a multiple-choice question.
8. Bounding boxes may be wrong and misled you. When answering questions about small objects in bounding boxes, you should zoom in on the object to see the details.
9. Segmentation and marking tool can help you better reason about the relationship between objects. Example use cases are spatial reasoning (e.g., left/right/above/below/on) and counting.
10. If a similar example is present, you should try to use the tools mentioned in the example, but you can use your reasoning to interpret the outputs of the tools.

Below are some examples of how to use the tools to solve the user requests. You can refer to them for help. You can also refer to the tool descriptions for more information.
# EXAMPLE: Answering a simple question about the image
# USER REQUEST #: <A image here> which city is this?
# USER IMAGE stored in image_1, as PIL image.
# RESULT #:
THOUGHT 0: The tools are not helpful for this task. I can directly answer the question.
ACTION 0: No action needed.
ANSWER: This looks like New York City. I can see the Empire State Building in the image. TERMINATE

# EXAMPLE: Detecting objects in the image
# USER REQUEST #: <A image here> is there a dog in the image? Options: (a) Yes (b) No
# USER IMAGE stored in image_1, as PIL image.
# RESULT #:
THOUGHT 0: Firstly, I need to detect dogs in the image.
ACTION 0: 
```python
image_1_dog_detection = generate_image("Detect all the dogs in the image and draw bounding boxes around them. Annotate the bounding boxes with labels.", image_1)
display(image_1_dog_detection)
```
OBSERVATION: Execution success. The output is as follows:
<the image outputs of the previous code is here.>
If you can get the answer, please reply with ANSWER: <your answer> and ends with TERMINATE. Otherwise, please generate the next THOUGHT and ACTION.
THOUGHT 1: A dog is detected by the tool. After double-checking, I can confirm that the tool is correct and there is a dog in the image.
ACTION 1: No action needed.
ANSWER: Yes, there is a dog in the image. So the answer is (a). TERMINATE

# EXAMPLE: Searching for small objects in the image
# USER REQUEST #: <A image here> What animal is drawn on that red signicade?
# USER IMAGE stored in image_1, as PIL image.
# RESULT #:
THOUGHT 0: I do not see a sign that is both red and has an animal drawn on it. I need to use the sliding_window_detection tool to find that sign and zoom in on it.
ACTION 0:
```python
possible_patches = sliding_window_detection(image_1, ["animal", "sign"])
for i, patch in enumerate(possible_patches):
    print(f"Patch {i+1}:")
    display(patch.annotated_image)
```
OBSERVATION: Execution success. The output is as follows:
Patch 1: 
<image output 1>
Patch 2:
<image output 2>
<More patches below...>
THOUGHT 1: Among the above patches, Patch 4 contains the animal drawn on the red signicade. An animal was detected. I will zoom in to double check if the detection is correct.
ACTION 1: 
```python
relevant_patch = possible_patches[3]
display(relevant_patch.annotated_image)
```
OBSERVATION: Execution success. The output is as follows:
<the image outputs of the previous code is here.>
THOUGHT 2: I can confirm that there is an animal drawn on the red signicade. I will zoom in to see what animal it is.
ACTION 2:
```python
animal_box = relevant_patch.boxes[2]
zoomed_patch = zoom_in_image_by_bbox(relevant_patch.original_image, animal_box)
display(zoomed_patch)
```
OBSERVATION: Execution success. The output is as follows:
<the image outputs of the previous code is here.>
THOUGHT 3: It can be seen that the animal drawn on the red signicade is a tiger.
ACTION 3: No action needed.
ANSWER: The animal drawn on the red signicade is a tiger. TERMINATE

# EXAMPLE: Reasoning about the depth of the objects     # As a Multimodal LLM, you might not be good at assessing the distance of objects from the camera, so it is good to generate a depth map to be sure.
# USER REQUEST #: <A image here> Which point is closer to the camera? Options: (A) Point A (B) Point B  
# USER IMAGE stored in image_1, as PIL image.
# RESULT #:
THOUGHT 0: I can use the generate_image tool to create a depth map for the image. This tool will provide a depth map where different colors represent different distances from the camera. 
I can also use the overlay_images tool to overlay the original image on the heat map to get a better sense of where point A and point B are on the depth map.
ACTION 0:
```python
image_1_depth = generate_image("Create a depth map for the image. Warmer colors represent closer objects.", image_1)
display(image_1_depth)
overlayed_image_1 = overlay_images(image_1_depth, image_1, alpha=0.3)
display(overlayed_image_1)
```
OBSERVATION: Execution success. The output is as follows:
<the image outputs of the previous code is here.>
THOUGHT 1: Let me crop into the area around points A and B, and look at the same points in the depth image. Then, I can compute the average temperature of the two regions to decide which one is closer to the camera.
ACTION 1: 
```python
# Define bounding boxes for A and B based on visual estimation from the original image
# A is roughly at x=0.25, y=0.6
# B is roughly at x=0.7, y=0.6
# I'll create crops around these areas.
box_a = [0.2, 0.5, 0.2, 0.2]
box_b = [0.6, 0.5, 0.2, 0.2]

crop_a = zoom_in_image_by_bbox(image_1, box_a)
crop_b = zoom_in_image_by_bbox(image_1, box_b)

crop_a_depth = zoom_in_image_by_bbox(image_1_depth, box_a)
crop_b_depth = zoom_in_image_by_bbox(image_1_depth, box_b)

display(crop_a_depth)
display(crop_b_depth)

average_temperature_crop_a = # compute the average temperature value of crop A
print(average_temperature_crop_a)

average_temperature_crop_b = # compute the average temperature value of crop b
print(average_temperature_crop_b)
```

The tool is helpful. From the overlayed image, I can see that the color in the circle under point B is warmer than that of point A, which means point B is closer to the camera.
ACTION 1: No action needed.
ANSWER: Based on the depth map, the circle under point B, which shows warmer colors, is closer to the camera than point A. So the answer is (B). TERMINATE

# EXAMPLE: Reasoning about the spatial relationship between objects
# USER REQUEST #: <A image here> What is on the left of the right laptop? Options: 
(A) Lamp 
(B) Chair 
(C) Plant 
(D) None of the above
# USER IMAGE stored in image_1, as PIL image.
# RESULT #:
THOUGHT 0: To help me better reason about the relationship between objects, I can create a segmentation mask for this image.
ACTION 0:
```python
image_1_segmented = generate_image("Create a segmentation mask segmenting all individual objects in this image. Label every segment with a number.", image_1)
overlayed_image_1 = overlay_images(image_1_segmented, image_1.annotated_image, alpha=0.5)
display(overlayed_image_1)
```
OBSERVATION: Execution success. The output is as follows:
<the image outputs of the previous code is here.>
THOUGHT 1: The tool is helpful. I can see the right laptop is labeled as 9. There is a lamp on its left, which is labeled as 12.
ACTION 1: No action needed.
ANSWER: On the left side of the right laptop, there is a lamp. So the answer is (A). TERMINATE

# EXAMPLE: # you must follow the plan in this example exactly for similar problems and use the prompt shown in the example. Compare the 'missing_piece' image to the options directly to determine your answer. Do NOT use any other tools (like crop or overlay) or code.
# USER REQUEST #: <An image here> <An image here> Given the object above. There is a missing piece in the white area. Which of the five pieces (A, B, C, D, or E) fits perfectly into the missing part of the object?
# USER IMAGES stored in image_1, image_2 as PIL image.
# RESULT #:
THOUGHT 0: First, I can use the generate image tool to remove the whitespace from the first image. Then, I can subtract the original image from it to obtain the piece that would complete the image.

ACTION 0:
```python
image_1_whitespace_removed = generate_image("Remove the white space from this image to reveal underlying pattern, don't change anything else", image_1)  # use this prompt as is. removing the whitespace works better than trying to fill in the whitespace.
missing_piece = subtract_images(image_1_whitespace_removed - image_1)
display(missing_piece)
```
OBSERVATION: Execution success. The output is as follows:
<the image outputs of the previous code is here.>
THOUGHT 1: Looking at the piece that would complete the image, as well as the options in image_2, the closest piece out of the five pieces (A, B, C, D, E) is piece X (where X is one of the options). Answer: (X). TERMINATE.

# EXAMPLE: # you must follow the plan in this example exactly for similar problems and use the prompt shown in the example. Compare the 'missing_piece' image to the options directly to determine your answer. Do NOT use any other tools (like crop or overlay) or code.
# USER REQUEST #: <An image here> <An image here> Given the object above. There is a missing piece in the white area. Which of the five pieces (A, B, C, D, or E) fits perfectly into the missing part of the object?
# USER IMAGES stored in image_1, image_2 as PIL image.
# RESULT #:
THOUGHT 0: First, I can use the generate image tool to remove the whitespace from the first image. Then, I can subtract the original image from it to obtain the piece that would complete the image.

ACTION 0:
```python
image_1_whitespace_removed = generate_image("Remove the white space from this image, don't change anything else", image_1)  # use this prompt as is. removing the whitespace works better than trying to fill in the whitespace.
missing_piece = subtract_images(image_1_whitespace_removed - image_1)
display(missing_piece)
```
OBSERVATION: Execution success. The output is as follows:
<the image outputs of the previous code is here.>
THOUGHT 1: The generate_image tool did not work well as the white jagged pattern is still there. Let me try again with the same prompt, as the outputs of the tool can be stochastic.

ACTION 1:
```python
image_1_whitespace_removed = generate_image("Remove the white space from this image, don't change anything else", image_1)  # use this prompt as is. removing the whitespace works better than trying to fill in the whitespace.
missing_piece = subtract_images(image_1_whitespace_removed - image_1)
display(missing_piece)
```

OBSERVATION: Execution success. The output is as follows:
<the image outputs of the previous code is here.>
THOUGHT 2: Looking at the piece that would complete the image, as well as the options in image_2, the closest piece out of the five pieces (A, B, C, D, E) is piece X (where X is one of the options). Answer: (X). TERMINATE.

# EXAMPLE: # Do NOT use the generate_image tool to count the points, as it is not good at counting objects. Also do not use it to highlight points in the overlapping region, count the points directly.
# You must use the generate_image tool to complete the image before counting the objects
# USER REQUEST #: <An image here> Your task is to count objects in the image. Assume the pattern of circles continues behind the black box. First, state what the pattern is, then give your final count. Put your answer inside \\boxed{}
# USER IMAGES stored in image_1 as PIL image.
# RESULT #:
THOUGHT 0: I can use the generate_image tool to remove the black square and complete the pattern.
ACTION 0: 
```python
image_1_completed = generate_image("Remove the black square from the image to complete the pattern.", image_1)
display(image_1_completed)
```

# EXAMPLE: # to predict collisions and trajectories of objects, you *must* use the generate image tool to plot the trajectory of the object
# USER REQUEST #: <An image here> Question: The aeroplane that the small red minivan will collide with if it moves forward is what color?
# USER IMAGES stored in image_1 as PIL image.
# RESULT #:
THOUGHT 0: I can use the generate_image tool to visualize the direction of movement of the small red minival.
ACTION 0: 
```python
image_1_completed = generate_image("Add a single straight line passing through the small red minivan. Only one line, passing right through the small red minivan. The line should be in the direction the red minivan is currently facing. Everything else remains the same.", image_1) # use this prompt as is (change the name of the object as per the quesiton)
display(image_1_completed)
```

OBSERVATION: Execution success. The output is as follows:
<the image outputs of the previous code is here.>
THOUGHT 1: The generated red line passes through...

# EXAMPLE:  # you must follow this strategy and this generate_image tool use for this task.
# USER REQUEST #: <An image here> <An image here> In the living room, the main door faces north; in which direction is the coffee table located in the dining room?Options: A: West, B: East, C: South, D: North
# USER IMAGES stored in image_1, image_2 as PIL image.
# RESULT #:
THOUGHT 0: I can use the generate_image tool to create a unified view of the scene using the two images. Then, I can use the 3D view to create a top-down blueprint of the scene.   # use this strategy for similar spatial reasoning tasks.
ACTION 0: 
```python
# the regions relevant to the question (e.g., main door, coffee table, dining room) should be mentioned in the prompt so that the model does not leave them out
# first create a 3d view of the scene, and use it to create a top-down blueprint of the scene. don't create the top-down view directly.
unified_3d_image = generate_image("use these two images to create a unified 3d view of the scene. the two images have overlap, use that to understand how they are related to each other. the accuracy in layout is the most important factor. Pay special attention to the main door, the coffee table and the dining room. create a 3d perspective image. accuracy and consistency with the input images is very important, do not hallucinate anything.", [image_1, image_2])    
display(unified_3d_image)
top_down_blueprint = generate_image("Based on the scene shown in this image, create a *top-down blueprint* of the scene. The accuracy of the layout is extremely important. Label the main door, the coffee table and the dining room. It is important that the orientation and layout of these objects is accurate. Triple check your layout before returning the output.", unified_3d_image)
display(top_down_blueprint)
```

OBSERVATION: Execution success. The output is as follows:
<the image outputs of the previous code is here.>
THOUGHT 1: I will use the *top_down_blueprint* to answer the question as it clearly shows the relative positions of various regions.
Based on the layout in the image, here is how we can figure out the direction:

Continue reasoning...

# EXAMPLE:  # you must follow this strategy and this generate_image tool use for this task.
# USER REQUEST #: <An image here> Instructions:\n\nTrace the lines to match each number (1, 2, 3, 4) to its corresponding letter (A, B, C, D).\n\nWrite down the letters in the ascending numerical order of their matches (from 1 to 4).\n\nNote that the lines do not intersect each other.\n\nExample: If 1 connects to B, 2 to C, 3 to A, and 4 to D, your final answer should be Answer: \"BCAD\"
# USER IMAGES stored in image_1 as PIL image.
# RESULT #:
THOUGHT 0: I can use the generate_image tool to convert the dashed lines into solid lines, which will make it easier to trace the paths without ambiguities.
ACTION 0: 
```python
solid_lines = generate_image("convert the dashed lines into solid lines. do not change the layout or anything else.", image_1)    
display(solid_lines)
```

OBSERVATION: Execution success. The output is as follows:
<the image outputs of the previous code is here.>
THOUGHT 1: Continue reasoning...
\end{tcblisting}
\clearpage

\subsection{Test-Time Scaling Selector Prompt}
\label{app:selector_prompt}

\texttt{\{prompt\}} is the agent's \texttt{generate\_image} prompt, $K$ the number of input images passed to it, and $N=10$ the number of candidates.

\begin{tcblisting}{enhanced,
  breakable,
  colback=white,
  colframe=black!80,
  boxrule=1pt,
  arc=2.5mm,
  left=2mm,
  right=2mm,
  top=1mm,
  bottom=1mm,
  listing only,
  listing options={
    basicstyle=\ttfamily\tiny,
    breaklines=true,
    breakatwhitespace=true,
    columns=fullflexible,
    keepspaces=true,
    showstringspaces=false
  }}
You are an image quality evaluator. Given the following generation prompt and candidate images,
select the image that BEST aligns with the prompt. Pay attention to the different requirements of the prompt and see whether the image meets the requirements.
When evaluating blueprints or 3d views of scenes, prioritize the accuracy of the layout, the orientation of various objects, etc. Factors like professionalism, labels, style, colors, and other minor details are not important. The placement of objects is very important. Especially pay attention to the objects mentioned in the prompt, their layout is crucial.

Original generation prompt: "{prompt}"

The generation was based on {K} input image(s) shown first.

[Input Image 1]
...
[Input Image K]

Below are {N} candidate generated images. Evaluate each one and select the BEST one.

Input Image 1: <image>
...
Input Image K: <image>

Candidate Image 1: <image>
...
Candidate Image N: <image>

Please analyze each candidate image and select the best one (1-{N}).

For each candidate, briefly evaluate:
1. How well the image matches the prompt requirements
2. Visual quality, coherence, and lack of artifacts
3. If input images were provided, how well the output preserves/transforms relevant elements

Provide your reasoning for each candidate, then state your final selection.

End your response with a line in this exact format:
SELECTED: <number>

For example, if candidate 3 is best, end with:
SELECTED: 3
\end{tcblisting}

%% file: files/related_work_extended.tex
\textbf{Visual reasoning with expert tools.} To go beyond purely textual chain-of-thought, a growing line of work equips Multimodal LLMs with visual tools that produce intermediate images the model can then reason over. For instance, \citet{hu2024visual,fu2025refocus} introduce tool-augmented agents that call perception or editing operators such as crop, zoom, detection, segmentation, or depth estimation, composing them over multiple steps. Recent zoom-centric systems such as Pixel Reasoner~\citep{wang2025pixelreasoner} and DeepEyes~\citep{zheng2025deepeyes} go further by training models to gather additional visual evidence through zoom-in or crop actions during reasoning, and similar tool-augmented frameworks have been developed for specialized domains~\citep{wang2025chartagenttir,mallis2025cadassistant,sharrock2025butterbench,zhao2025pyvision}. These systems remain bounded, however, by the tools available to them: each function performs a narrow, pre-specified transformation, so the model can only carry out manipulations that have been implemented in advance. Open-ended transformations such as removing an arbitrary occluder, drawing a hypothetical collision trajectory, or synthesizing a counterfactual view are out of reach unless a dedicated tool has already been built and the model is instructed to invoke it. In contrast, \ours\ replaces this fixed toolset with a single instruction-following generative model that can perform a wider range of visual transformations expressed in natural language, removing the need to engineer a new tool for each new visual operation.

\textbf{Iterative refinement of generated images.} A separate line of work interleaves language-based reasoning with repeated image synthesis, focusing on improving the final image itself rather than to solve a downstream reasoning task. \citet{yang2024idea2img} use GPT-4V to iteratively probe a text-to-image model, propose revised prompts, and select among drafts. \citet{guo2025can} verify and reinforce image generation step by step using specialized reward models, \citet{khan2025tir} have a pretrained MLLM inspect each generated image against the user's prompt and rewrite the prompt for the next synthesis round, and \citet{wan2025maestro} orchestrate multi-agent critique with tournament-based selection between iterations. In all of these settings, the focus is on the outputting an image that follows as much as possible the instructions of the user, whereas in our setting the image is an intermediate artifact and the goal is to answer a downstream question.

\textbf{Reasoning with generative visual models.} A growing line of work explores using image generation to produce intermediate visual artifacts within a reasoning process. \citet{li2025imaginereasoningspacemultimodal} and Visual Planning~\citep{xu2026visualplanningletsthink} reason entirely through sequences of generated images, without textual chain-of-thought, and target individual narrow domains such as maze navigation and embodied planning respectively. DiffThinker~\citep{he2025diffthinkergenerativemultimodalreasoning} similarly reformulates reasoning as a native image-to-image task using a diffusion transformer, and is also evaluated on a single domain of planning and spatial configuration problems. ThinkMorph~\citep{gu2025thinkmorph} interleaves image generation with textual chain-of-thought, but the visual operations it considers, such as zooming and overlaying, overlap substantially with what fixed specialist tools can already provide, not demonstrating the diversity and flexibility that powerful image generators offer. 
Complementing these methods, \citet{zeller2026mentisoculi} introduce a benchmark that probes the limits of mental imagery in multimodal models. 
A related line of work keeps the visual chain of thought internal, decoding continuous latent tokens rather than pixels. Each requires specialized training of the reasoning model, such as multi-stage distillation with reinforcement learning~\citep{yang2026machine} or distillation from vision experts~\citep{qin2025chain}. Their scope is also limited: some are evaluated only on narrow domains like maze navigation and spatial planning~\citep{zhang2025latent,yang2026machine}, while others operate broadly but have not been shown to go beyond what specialists such as depth estimation or grounding already provide~\citep{qin2025chain,li2026latent}. Finally, because their visual reasoning happens in latent space, the intermediate steps are not interpretable. In contrast, \ours\ is training-free and modular, and carries out its visual reasoning in pixel space, so every intermediate step is an explicit image that humans can directly inspect. Separately, recent work probes whether video generation models can serve as zero-shot visual reasoners by leveraging their learned world dynamics~\citep{wiedemer2025videomodels}; however, the generated video there is the final output being evaluated, rather than an intermediate artifact within a broader reasoning loop. In contrast, we study a training-free framework that interleaves text and image generation, evaluated across six diverse tasks with the same setup.

%% file: tables/results_cost.tex
\begin{table*}[!htbp]
  \centering
  \small
  \resizebox{\linewidth}{!}{
  \begin{tabular}{@{}l cccccc cc cc@{}}
  \toprule
  & \multicolumn{6}{c}{\textbf{MLLM reasoning}} & \multicolumn{2}{c}{\textbf{Image generation}} & & \\
  \cmidrule(lr){2-7} \cmidrule(lr){8-9}
  \textbf{Method}
    & \shortstack{\textbf{In} \\ {\scriptsize (tok.)}}
    & \shortstack{\textbf{Out} \\ {\scriptsize (tok.)}}
    & \shortstack{\textbf{Total} \\ {\scriptsize (tok.)}}
    & \shortstack{\textbf{Calls}}
    & \shortstack{\textbf{Images}}
    & \shortstack{\textbf{Cost} \\ {\scriptsize (USD)}}
    & \shortstack{\textbf{Tokens}}
    & \shortstack{\textbf{Cost} \\ {\scriptsize (USD)}}
    & \shortstack{\textbf{Total cost} \\ {\scriptsize (USD)}}
    & \shortstack{\textbf{Median time} \\ {\scriptsize (s)}} \\
  \midrule
    No Tools    & $1{,}700$  & $600$ & $2{,}300$  & $1.00$ & $0$   & $0.01$ & --         & --      & $0.01$ & $38$ \\
    Sketchpad   & $21{,}000$ & $400$ & $21{,}500$ & $1.83$ & $1.3$ & $0.04$ & --         & --      & $0.04$ & $52$ \\
    \rowcolor{blue!10} \ours & $19{,}700$ & $350$ & $20{,}000$ & $1.79$ & $1.2$ & $0.04$ & $1{,}300$ & $0.12$ & $0.16$ & $76$ \\
  \bottomrule
  \end{tabular}
  }
  \caption{
  \textbf{Per-instance inference cost of \ours\ compared to baselines}, averaged over all task instances with Gemini-3.1-Pro as the MLLM and Nano-Banana-Pro as the visual generative tool.
  \ours\ and Sketchpad are \emph{matched} on every quantity that reflects reasoning effort---MLLM input, output, and total tokens, number of MLLM calls, number of intermediate images, and the resulting MLLM cost (\$$0.04$ for both)---and therefore spend the same effort on reasoning.
  The two methods differ only in the cost and latency of image generation (Image generation columns): Sketchpad's intermediate images come from specialist vision models that incur no generation-token cost, whereas \ours\ synthesizes them with a frontier image model.
  We expect this gap to narrow as generative models become cheaper and faster.
  Costs are in USD per task instance; the one-time cost of strategy discovery is reported separately in Appendix~\ref{app:strategy_prompts}.
  }
  \label{tab:cost}
\end{table*}

%% file: tables/results_open_gen_qwen.tex
\begin{table*}[!htbp]
  \centering
  \small
  \resizebox{\linewidth}{!}{
  \begin{tabular}{@{}l cccccc@{}}
  \toprule
  \textbf{Image Gen.\ Model} & \shortstack{\textbf{Puzzle $\uparrow$} \\ {\scriptsize (MIRA)}} & \shortstack{\textbf{Occ.\ Count $\downarrow$} \\ {\scriptsize (CAPTURe)}} & \shortstack{\textbf{Collision $\uparrow$} \\ {\scriptsize (Spatial457)}} & \shortstack{\textbf{Spatial $\uparrow$} \\ {\scriptsize (MMSI)}} & \shortstack{\textbf{Tracing $\uparrow$} \\ {\scriptsize (Ours)}} & \shortstack{\textbf{Depth $\uparrow$} \\ {\scriptsize (BLINK)}} \\
  \midrule
    No Tools              & $34.6$  &  $12.1$  &  $51.7$  &  $42.0$  & $35.7$  &  $85.5$ \\
    Nano-Banana-Pro       & $\boldsymbol{46.2}$  &  $6.4$  & $\boldsymbol{61.3}$  &  $\textbf{48.0}$  &  $\boldsymbol{76.0}$  &  $\boldsymbol{90.1}$ \\
    FLUX.2 [dev]          & $44.2$  & $\boldsymbol{6.1}$  & $45.5$  & $38.0$  & $66.0$  & $85.5$ \\
    Qwen-Image-Edit-2511  & $32.1$  & $8.8$  & $50.0$  & $42.3$  & $48.0$  & $80.3$ \\
  \bottomrule
  \end{tabular}
  }
  \caption{
  \textbf{Effect of the image generation model on \ours\ performance, with Qwen-3.5-27B as the MLLM.}
  This repeats the comparison of Table~\ref{tab:open_gen} with an open-weights MLLM in place of Gemini-3.1-Pro, giving a fully open-weights configuration in the two open-generator rows.
  The overall picture matches the Gemini setting: Nano-Banana-Pro is best on five of six tasks. Both open-weights generators improve over the No Tools baseline on occlusion counting and path tracing, and FLUX.2 [dev] also on puzzle completion, where it comes close to Nano-Banana-Pro.
  The exceptions are collision prediction, spatial reasoning, and depth, where both open generators fall to or below No Tools, mirroring their weakness on spatially precise transformations in Table~\ref{tab:open_gen}.
  }
  \label{tab:open_gen_qwen}
\end{table*}

%% file: tables/results_no_generate.tex
\begin{table*}[!htbp]
  \centering
  \small
  \resizebox{\linewidth}{!}{
  \begin{tabular}{@{}l cccccc@{}}
  \toprule
  \textbf{Approach} & \shortstack{\textbf{Puzzle $\uparrow$} \\ {\scriptsize (MIRA)}} & \shortstack{\textbf{Occ.\ Count $\downarrow$} \\ {\scriptsize (CAPTURe)}} & \shortstack{\textbf{Collision $\uparrow$} \\ {\scriptsize (Spatial457)}} & \shortstack{\textbf{Spatial $\uparrow$} \\ {\scriptsize (MMSI)}} & \shortstack{\textbf{Tracing $\uparrow$} \\ {\scriptsize (Ours)}} & \shortstack{\textbf{Depth $\uparrow$} \\ {\scriptsize (BLINK)}} \\
  \midrule
    \rowcolor{blue!10} \ours                                & $\textbf{42.3}_{\pm 2.2}$  & $\textbf{7.1}_{\pm 0.6}$  & $\textbf{69.7}_{\pm 1.2}$  & $\textbf{59.0}_{\pm 0.6}$  &  $\textbf{91.0}_{\pm 1.7}$  & $\textbf{94.6}_{\pm 1.5}$ \\
    \ours\ w/o \texttt{generate\_image}  & $36.5_{\pm 5.8}$  & $7.7_{\pm 0.4}$  & $65.0_{\pm 2.0}$  & $54.5_{\pm 0.2}$  & $77.5_{\pm 2.5}$  & $92.7_{\pm 0.0}$ \\
  \bottomrule
  \end{tabular}
  }
  \caption{
  \textbf{Effect of removing the generative tool from \ours}, with Gemini-3.1-Pro as the MLLM.
  The ablated variant replaces \texttt{generate\_image} with a no-op that returns the input image unchanged, keeping everything else identical: the same prompts, the same tool definitions, the same Python execution environment, and the same programmatic image utilities such as \texttt{crop}, \texttt{overlay}, and \texttt{subtract\_images}.
  Because both rows share identical prompts and scaffolding, this comparison isolates the generative capability as the single variable, unlike the comparison against No Tools and Visual Sketchpad in Table~\ref{tab:main}, which also differ in their prompts and agent scaffolds.
  Performance drops on all six tasks.
  Test-time scaling selects among sampled generations and is therefore inapplicable to the ablated variant, whose samples are all identical to the input.
  }
  \label{tab:no_generate}
\end{table*}